\documentclass[10pt,twocolumn,letterpaper]{article}

\usepackage[pagenumbers]{wacv} 

\usepackage[accsupp]{axessibility}

\usepackage{multirow}
\usepackage{subcaption}
\usepackage[dvipsnames]{xcolor}
\usepackage{bm}

\usepackage{colortbl}
\usepackage{array}
\newcolumntype{M}{@{\hspace{.3em}}c@{\hspace{.3em}}}
\definecolor{wacvblue}{rgb}{0.21,0.49,0.74}
\usepackage[pagebackref,breaklinks,colorlinks,allcolors=wacvblue]{hyperref}

\def\wacvPaperID{936} 
\def\confName{WACV}
\def\confYear{2026}

\title{DNA: Dual-branch Network with Adaptation for \\ Open-Set Online Handwriting Generation}

\author{
Tsai-Ling Huang\thanks{Equal contribution \qquad $^{\dagger}$Corresponding author} \qquad
Nhat-Tuong Do-Tran$^{\ast}$ \qquad
Ngoc-Hoang-Lam Le$^{\ast}$ \\
Hong-Han Shuai \qquad \qquad
Ching-Chun Huang$^{\dagger}$ \\
National Yang Ming Chiao Tung University \\
{\tt\footnotesize \{christina.ii12, tuongdotn.ee12, lengochoanglam.ee12, hhshuai, chingchun\}@nycu.edu.tw}
}

\begin{document}
\maketitle
\begin{abstract}
Online handwriting generation (OHG) enhances handwriting recognition models by synthesizing diverse, human-like samples. However, existing OHG methods struggle to generate unseen characters, particularly in glyph-based languages like Chinese, limiting their real-world applicability. In this paper, we introduce our method for OHG, where the writer’s style and the characters generated during testing are unseen during training. To tackle this challenge, we propose a \underline{D}ual-branch \underline{N}etwork with \underline{A}daptation (DNA), which comprises an adaptive style branch and an adaptive content branch. The style branch learns stroke attributes such as writing direction, spacing, placement, and flow to generate realistic handwriting. Meanwhile, the content branch is designed to generalize effectively to unseen characters by decomposing character content into structural information and texture details, extracted via local and global encoders, respectively. 
Extensive experiments demonstrate that our DNA model is well-suited for the unseen OHG setting, achieving state-of-the-art performance.
\vspace{-.5em}

\end{abstract}    
\section{Introduction}
\label{sec:intro}

\begin{figure}[!t]
  \centering
    \includegraphics[width=\columnwidth]{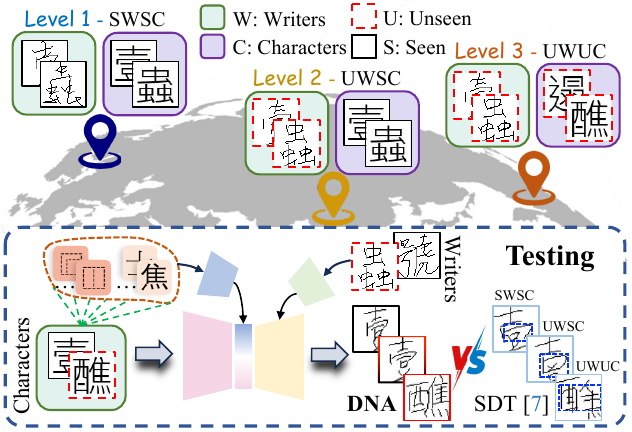}
    \vspace{-1.5em}
    \caption{Illustration of the online handwriting generation setting. Our proposed DNA model is trained under the SWSC (Seen-Writer-Seen-Characters) Level 1 setting. During testing, we evaluate DNA’s ability to generate online handwriting characters across two other levels: UWSC (Unseen-Writer-Seen-Characters) and UWUC (Unseen-Writer-Unseen-Characters). Additionally, we compare DNA’s performance with the state-of-the-art UWSC method, SDT \cite{SDT}. The blue squares highlight incorrect strokes generated by SDT, particularly at the UWUC Level 3 setting.}
    \vspace{-1.5em}
    \label{fig:intro}
\end{figure}


While handwriting recognition networks \cite{Metahtr, HLS, lee2025improving} have achieved promising results, real-world deployment remains challenging due to the high cost and time required for collecting large-scale annotated datasets; moreover, even with extensive data, these models may struggle to generalize to unseen test samples \cite{Metahtr, ECB}. To address this limitation, zero-shot approaches \cite{SideNet, ao2024prototype, chen2021zero} aim to improve generalization by leveraging semantic embeddings and learning relationships between visual features and unseen style attributes. Meanwhile, few-shot methods \cite{wang2019radical, FFD-Augmentor, samuel2022offline} enhance recognition by adapting to new handwriting styles using a small set of labeled examples. However, both zero-shot and few-shot recognition primarily focus on designing better learning models, often leading to suboptimal performance on unseen data. This raises a key question: \textbf{Can recognition generalization be further improved through the synthesis of more effective training data?}
A common solution is data augmentation \cite{shijie2017research, Ricap, takahashi2019data}, which increases dataset diversity using techniques such as flipping, cropping, and more advanced transformations \cite{chen2024fine, de2024data, wigington2017data}. However, traditional data augmentation cannot generate new words or capture handwriting variations from unseen writers. To overcome this, recent handwriting generation approaches \cite{zhao2024cross, Fontdiffuser, CHWmaster, SDT} offer an alternative by synthesizing diverse character sets that closely mimic human-like handwriting, thereby improving recognition generalization.
As illustrated in \cref{fig:intro}, the handwriting generation task can be categorized into level (\textbf{1}) SWSC (Seen-Writer-Seen-Characters): Both the writers and characters are observed during training and testing. Level (\textbf{2}) UWSC (Unseen-Writer-Seen-Characters): The model generates handwriting for seen characters using the styles of unseen writers. Level (\textbf{3}) UWUC (Unseen-Writer-Unseen-Characters): The most challenging setting, where the model must generate unseen characters in the style of unseen writers during testing.


Furthermore, handwriting generation can be categorized into offline \cite{Dg-font, Cf-font, Fontdiffuser} and online \cite{WriteLikeYou, SDT, DeepCalliFont} methods. Offline generation can synthesize handwriting that imitates a particular writer’s style. Despite significant progress at level (\textbf{1}), these offline methods directly generate static character images, making it difficult to capture the complex and dynamic variations of handwriting styles naturally. As a result, they often suffer from stroke connection errors and structural inconsistencies in the generated handwriting. To address this issue, online handwriting generation methods represent glyphs as ordered point sequences, capturing stroke trajectories more effectively. For instance, state-of-the-art (SoTA) methods \cite{SDT, WriteLikeYou} introduce a disentangled style representation, separating glyph attributes from writer-specific features. This enables the model to extract writing style from limited reference samples from the target writer. 
However, existing online methods \cite{SDT, DeepCalliFont, Elegantly-Written} extract character content from input images, failing to fully preserve crucial structural details. As shown in \cref{fig:intro} (blue squares), SDT struggles to accurately generate unseen characters' structures at level (\textbf{2}) and (\textbf{3}) settings, leading to writing stroke errors. A straightforward solution might be to apply diffusion models \cite{Fontdiffuser, Diff-font} for handwriting synthesis. However, diffusion models are computationally intensive and primarily follow a map-to-map translation approach, which does not explicitly model the dynamics of stroke trajectories. As a result, generated handwriting often suffers from stroke connection errors. These observations highlight the challenges of the practical setting: unseen \underline{O}nline \underline{H}andwriting \underline{G}eneration (OHG), which aims to dynamically generate handwriting for unseen characters in the styles of unseen writers (level (\textbf{3})).

To match the practical UWUC scenario, we propose a novel \underline{D}ual-branch \underline{N}etwork with \underline{A}daptation (DNA), which consists of a content branch and a style branch. Our DNA extracts fine-grained writing styles from reference samples and integrates them with character content from another content sample to generate new handwriting samples under the unseen OHG setting. Specifically, DNA is trained solely in the SWSC setting, yet it can mimic the handwriting of unseen individuals and generate any unseen characters under UWSC and UWUC settings during testing. To synthesize handwriting in unseen styles, DNA introduces an \textbf{adaptive style branch} that disentangles writing styles from reference handwriting samples. Unlike prior style-disentanglement methods \cite{DeepCalliFont, SDT, WriteLikeYou}, DNA incorporates a novel spacing loss ($\mathcal{L}_{sp}$) to preserve stroke spacing.
On the other hand, the \textbf{adaptive content branch} disentangles character text into structural components and texture through a local content encoder and a global texture encoder. The local component encoder extracts structural information by decomposing characters into word structures and components $\big($e.g., \includegraphics[width=0.37cm, height=0.3cm]{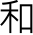} consists of structure \includegraphics[width=0.3cm, height=0.3cm]{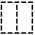} and components \includegraphics[width=0.2cm, height=0.3cm]{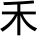}, \includegraphics[width=0.17cm, height=0.3cm]{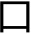}$\big)$. Meanwhile, the global content encoder captures high-level representations to model texture information. To further enhance content generalization, we introduce a content loss ($\mathcal{L}_{ct}$), including novel character loss ($\mathcal{L}_{ch}$) and a decomposition loss ($\mathcal{L}_{de}$), which guide interactions between the local content encoder and the global texture encoder during training. By effectively complementing structural and texture features, DNA can robustly generate characters, even those unseen in the training data. We summarize the key contributions of DNA as follows:
\begin{enumerate}
    \item We introduce the novel unseen \underline{O}nline \underline{H}andwriting \underline{G}eneration (OHG) task and propose a \underline{D}ual-branch \underline{N}etwork with \underline{A}daptation (DNA) to synthesize handwriting for both unseen writers and characters.
    \item We incorporate a spacing loss ($\mathcal{L}_{sp}$) to enhance the adaptive style branch, ensuring better alignment between generated strokes and the ground truth, thereby making the online handwriting more natural and fluid. 
    \item An adaptive content branch is designed to capture both local and global information by complementing structural and texture features, improving overall content representation. Additionally, we introduce two novel losses — character loss ($\mathcal{L}_{ch}$) and decomposition loss ($\mathcal{L}_{de}$) — to improve the generalization of this content branch.
    \item Quantitative and qualitative evaluations demonstrate that DNA achieves SoTA performance on Traditional Chinese and Japanese handwriting datasets.
\end{enumerate}





\section{Related Works}
\label{sec:2_related}

\begin{figure*}[!t]
  \centering
   \includegraphics[width=0.75\textwidth]{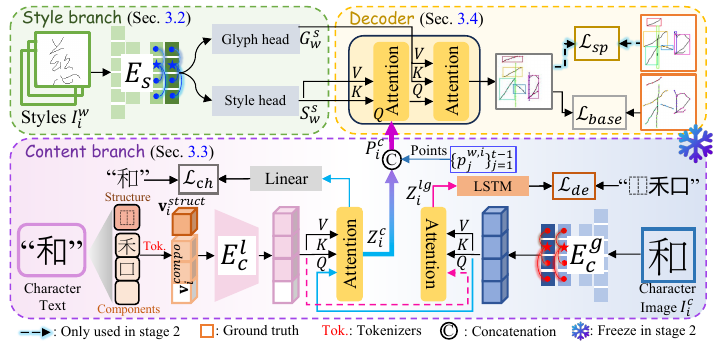}
   \vspace{-1em}
   \caption{Illustration of our proposed DNA. The style branch employs a style encoder $E_s$, followed by style and glyph heads, to extract style features $S_w^s$ and $G_w^s$. The content branch uses a local encoder $E^l_c$ for structural features and a global encoder $E^g_c$ for texture features. Structural features enhance texture features to form the enriched character content embedding $Z^{c}_i$. Along with previous point states $\{p^{w,i}_j\}_{j=1}^{t-1}$, this forms the point sequence $P^c_i$. The decoder then integrates $P^c_i$ with $S_w^s$ and $G_w^s$ to generate the online handwriting trajectory. Note that the spacing loss $\mathcal{L}_{sp}$ is applied only in the second training stage to ensure fluid handwriting.}
   \label{fig:method} 
   \vspace{-1.em}
\end{figure*}

\noindent \textbf{Handwriting Generation Setting.} 
In real-world scenarios, generating handwriting in numerous unseen writer styles (UW) and unseen characters (UC) is crucial, particularly for glyph-based handwriting synthesis. However, existing handwriting generation models \cite{Cf-font, park2021few, tang2022few} struggle in the UWUC setting. Diffusion models \cite{Diff-font, Fontdiffuser} address the UW setting by leveraging advanced image generation capabilities. However, they primarily learn style from images, making it difficult to capture handwriting stroke trajectories for fluid handwriting generation. As a result, online methods \cite{WriteLikeYou, DeepCalliFont} have gained attention, as they can trace writer-specific stroke trajectories. Notably, SDT \cite{SDT} employs a style disentanglement-based learning strategy to better model stroke sequences, enabling the replication of both seen and unseen writers. Despite these advances, existing methods often overlook the vast number of glyph characters. Consequently, when generating unseen characters, their performance remains inconsistent. This necessity for robust UWUC sample generation then raises the online handwriting generation (OHG) challenge.


\noindent \textbf{Online Handwriting Generation.} Recently, numerous methods have been developed for online handwriting generation, enabling models to dynamically learn character writing with improved generalization to human-like handwriting. Early RNN-based approaches \cite{FontRNN, zhang2017drawing} focused on generating readable Chinese characters in specified classes, emphasizing calligraphy style over content. However, these methods were limited to generating styles present in the training set. Subsequent works, such as DeepImitator \cite{Deep-imitator}, WriteLikeYou \cite{WriteLikeYou}, and SDT \cite{SDT}, introduced the ability to generate characters in unseen styles using only a few reference samples. DeepImitator employs a CNN-based style encoder to extract local style features from handwritten character images. WriteLikeYou leverages multiple attention mechanisms to capture both global content and style. SDT applies a disentanglement strategy, separating reference images into glyph and writer style. Despite advances in online handwriting generation for unseen writers, previous works have ignored the stroke spacing control and unseen character generation. In this paper, we aim to preserve stroke spacing for more realistic handwriting while accurately capturing the writer’s style. Additionally, we propose an adaptive content branch to generate fine-grained handwriting characters under the \textcolor{black}{unseen OHG} setting.

\section{Method}
\label{sec:3_method}
In this section, we present the design of our proposed method, DNA, a framework developed for the \textcolor{black}{unseen OHG} problem, as illustrated in \cref{fig:method}. Specifically, we define the \textcolor{black}{unseen OHG} challenge (\cref{sec:problem}), elaborate on our adaptive style branch (\cref{sec:style}) and adaptive content branch (\cref{sec:content}), and describe the decoder architecture (\cref{sec:decoder}). Additionally, we detail our training strategy (\cref{sec:training}).

\begin{figure*}[!ht]
  \centering
  \begin{subfigure}{0.78\linewidth}
  \centering
    \includegraphics[width=0.95\linewidth]{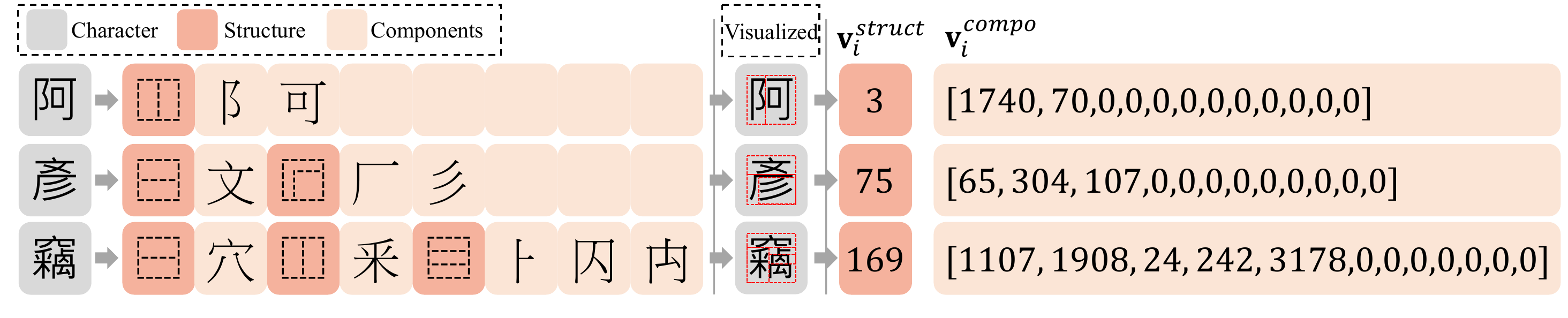}
    \caption{Examples of decomposition of Chinese characters and their corresponding encoding vectors, $\mathbf{v}^{struct}_{i}$ and $\mathbf{v}^{compo}_{i}$.}
    \label{fig:decompo_vector}
  \end{subfigure}
  \hfill
  \begin{subfigure}{0.18\linewidth}
  \centering
    \includegraphics[width=0.8\linewidth]{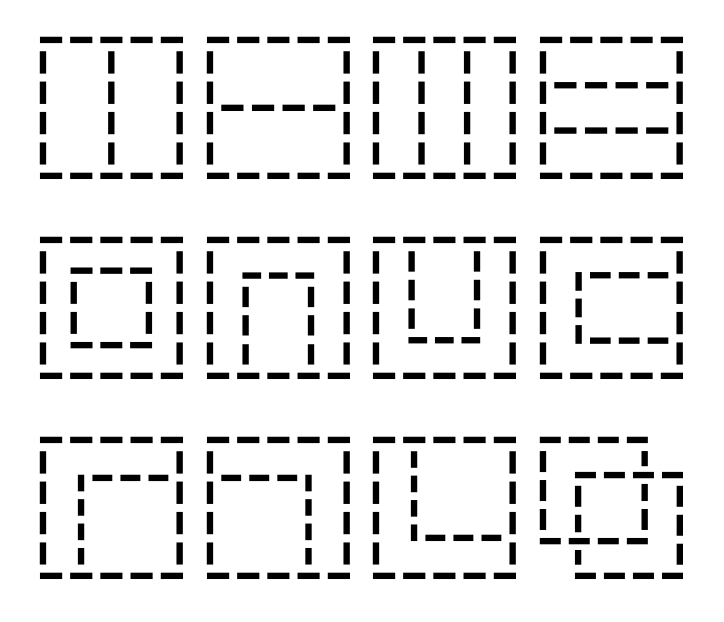}
    \caption{$12$ fundamental structures of characters.}
    \label{fig:structure}
  \end{subfigure}
  \vspace{-0.5em}
  \caption{Schematic diagram illustrating the structural and component decomposition of Chinese characters.} 
  \vspace{-1.em}
  \label{fig:compo_and_struct}
\end{figure*}

\subsection{Problem Definition} \label{sec:problem}
In the \textcolor{black}{unseen OHG} setting, a generative network $f_{\theta}(\cdot, \cdot)$ is trained using a style dataset consisting of handwriting images, denoted as $X_s = \{x^s_w\}_{w=1}^{N_w}$, where $N_w$ represents the number of writers. Each writer’s set, $x^s_w = \{I^{w}_0, I^{w}_1, ..., I^{w}_{N_s-1}\}$, contains $N_s$ style images ($I^{w}_i$). A content dataset of printed images is also provided for training, denoted as $X_c = \{(\mathbf{v}^c_i, I^c_i)\}_{i=1}^{N_c}$, where $N_c$ is the number of seen characters. Each character’s content composition is represented as $\mathbf{v}^c_i = \{\mathbf{v}^{struct}_{i}, \mathbf{v}^{compo}_{i}\}$, where $\mathbf{v}^{struct}_{i}$ and $\mathbf{v}^{compo}_{i}$ correspond to its structural and component vectors, respectively. \textcolor{black}{These vectors for both Chinese and Japanese characters are tokenized based on the IDS rules\footnote{\href{https://github.com/cjkvi/cjkvi-ids}{https://github.com/cjkvi/cjkvi-ids}}. As illustrated in \cref{fig:compo_and_struct}, Chinese characters comprise $12$ fundamental structures and $9,813$ basic components. In the Japanese setting, besides the same fundamental structures as Chinese, characters contain 1602 basic components.} Note that each content image, $I^c_i \in \mathbb{R}^{3\times H\times W}$, is rendered using the Microsoft JhengHei font. The ground truth for online handwriting characters of the $w$-th writer and the $i$-th character is represented by a point set, $Y_{w,i} = \{p_j^{w,i}\}_{j=1}^{N_p}$, where $N_p$ is the number of points. Each point $p_j^{w,i}=\{\Delta {x}_j, \Delta {y}_j, {m}^1_j, {m}^2_j, {m}^3_j\}$ includes two point offsets and three pen states (pen-down, pen-up, and pen-end).

The goal of \textcolor{black}{unseen OHG} is to train the model $f_{\theta}(\cdot, \cdot)$ to generate novel handwriting samples that preserve the writing style of the reference samples from the style branch while maintaining the content specified by the inputs of the content branch. During training, the model is provided with a seen-writer, seen-character (SWSC) dataset, denoted as $\mathcal{D}^{trn}=\{(X^{seen}_s, X^{seen}_c)\}$. At the testing stage, we evaluate at the other two levels using the following datasets:

\begin{enumerate}
    \item \textbf{Unseen writer, seen character (UWSC):} The model is tested on an unseen writer style, denoted as $X^{un}_s$, while the set of characters remains unchanged, forming $\mathcal{D}^{unw} = \{(X^{un}_s, X^{seen}_c)\}$.
    \item \textbf{Unseen writer, unseen character (UWUC):} Both the writer styles and characters are absent from the training dataset, resulting in $\mathcal{D}^{un}=\{(X^{un}_s, X^{un}_c)\}$.
\end{enumerate}

\subsection{Adaptive Style Branch} \label{sec:style}
As shown in the \textcolor{ForestGreen}{green} section of \cref{fig:method}, our DNA framework incorporates an adaptive style branch, which consists of three key modules: a style encoder $E_s$, a style head $h_s$, and a glyph head $h_g$, with a set of learnable parameters $\theta_s = \{\theta_{E_s}, \theta_{h_s}, \theta_{h_g}\}$. The style encoder $E_s$ first selects a mini-batch of style images $x^s_w$ of the $w$-th writer. These images are then processed by $E_s$ to extract style patterns, represented as $Z^s_w = E_s(x^s_w)$. This results in feature maps of dimensions $\mathbb{R}^{N_s \times d\times d'}$ from the input space $\mathbb{R}^{N_s \times 3 \times H \times W}$, where $d$ and $d'$ denote the embedding dimensions, and $H$ and $W$ represent the height and width of the style images, respectively. Next, the extracted style patterns $Z^s_w$ are further processed by the style head $h_s$ and the glyph head $h_g$ for feature decomposition. This process produces writer-specific style representations $S^s_w = h_s(Z^s_w) \in \mathbb{R}^{N_s \times d \times d'}$ and glyph-specific style representations $G^s_w = h_g(Z^s_w) \in \mathbb{R}^{N_s \times d \times d'}$. These representations, $S^s_w$ and $G^s_w$, are later utilized in the decoder for handwriting generation, as detailed in \cref{sec:decoder}.

\subsection{Adaptive Content Branch} \label{sec:content}
To enhance generalization for unseen characters, we propose an adaptive content branch comprising two parallel encoders: a local encoder $E^l_{c}(\cdot, \theta_{E^l_{c}})$ and a global encoder $E^g_{c}(\cdot, \theta_{E^g_{c}})$, represented by the \textcolor{CarnationPink}{pink}/left and \textcolor{blue}{blue}/right parts in \cref{fig:method}, respectively. Furthermore, we design a fusion decoder to integrate local and global features effectively.

\noindent \textbf{Local encoder.} To extract a character’s structural representation, we first apply an embedding layer $f_e$ to learn attributes from the structure and component of the ``character text'' input. We then leverage an attention mechanism \cite{attention} of a transformer $f_t$ to capture the relationship between the character's structure and its components. Formally, our local encoder is defined as $E^l_{c}=\{f_e, f_t\}$ and produces a local representation $Z^{lc}_i=E^l_{c}(\mathbf{v}^c_i) \in \mathbb{R}^{d^l\times d'}$. More specifically, each decomposition vector $\mathbf{v}^c_i = \{\mathbf{v}^{struct}_{i}, \mathbf{v}^{compo}_{i}\}$ is passed through $f_e$ individually, and then concatenated before being fed into the $f_t$ as follows:
\begin{equation}
Z^{lc}_i=f_t\big( \bm{[}f_e(\mathbf{v}^{struct}_{i}),f_e(\mathbf{v}^{compo}_{i})\bm{]}\big),
\label{eq:1}
\end{equation}
\noindent where $\bm{[}\cdot, \cdot \bm{]}$ is the concatenation operation, ensuring $f_t$ jointly considers structural and component features.

\noindent \textbf{Global encoder.} To capture the broader textual content representation of a character, we construct the global encoder $E^g_{c}(\cdot, \theta_{E^g_{c}})$ as a sequential combination of a Convolutional Neural Network (CNN) \cite{CNN} and a Vision Transformer (ViT) \cite{ViT}. The CNN first extracts regional features from the input character image $I^c_i$. To enhance global understanding beyond these regional features, ViT captures global texture information of the character. In summary, the global encoder works as follows:
\begin{equation}
Z^{gc}_{i} = E^g_{c}(I^c_i),
\label{eq:2}
\end{equation}
where $Z^{gc}_{i} \in \mathbb{R}^{d^g \times d'}$ represents the global texture representation of the input character image $I^c_i$.


\noindent \textbf{Cross-Attention Fusion.} To enhance texture representations extracted by the global encoder with local structural information from the local encoder, thereby producing the fused and enriched character content embedding $Z^{c}_i$, we employ a Cross-Attention (CA) mechanism. The CA operation, denoted as CA$(x, y)$, is computed using queries (${x}$), keys (${y}$), and values (${y}$).
The fusion representation is then computed by multi-head cross-attention as follows:
\begin{equation}
Z^{c}_i = Avg\big(\text{CA}({\bar{Z}}^{gc}_i, {\bar{Z}}^{lc}_i) + Z^{gc}_i\big),
\label{eq:4}
\end{equation}
where ${\bar{Z}}^{gc}_i = ln(Z^{gc}_i)$ and ${\bar{Z}}^{lc}_i = ln(Z^{lc}_i)$ represent the layer-normalized versions of the global representation $Z^{gc}_i$ and local representation $Z^{lc}_i$, respectively. $ln$ denotes layer normalization. $Avg$ represents the average operation applied along the first dimension, i.e. $d^g$. Finally, $Z^{c}_i \in \mathbb{R}^{1 \times d'}$, the fused feature, is then combined with $S^s_w$ and $G^s_w$ (as detailed in \cref{sec:style}) to synthesize online handwriting sequences using the decoder (\cref{sec:decoder}). 

On the other hand, we introduce two auxiliary tasks to enhance the generalization ability of the content branch, ensuring that the local and global content extractors complement each other. First, we employ the character recognition loss $\mathcal{L}_{ch}$ to supervise the content branch, enabling the final fused content features $Z^{c}_i$ for character recognition. Second, we refine local features $Z^{lc}_i$ by aggregating information from global features $Z^{gc}_i$ through cross-attention:
\begin{equation}
Z^{lg}_i = Avg\big(\text{CA}({\bar{Z}}^{lc}_i, {\bar{Z}}^{gc}_i) + Z^{lc}_i\big).
\end{equation}
To ensure that the content branch preserves structural information, the refined features $Z^{lg}_i$ are processed by an LSTM \cite{LSTM} to extract component sequences. The process is supervised by the decomposition loss $\mathcal{L}_{de}$, as detailed in \cref{sec:training}.

\subsection{Decoder} \label{sec:decoder}
After extracting the style features $S^s_w$ and $G^s_w$ for the $w$-th writer and the detailed content features $Z^c_i$ for the $i$-th character, we follow \cite{SDT, WriteLikeYou} and employ a decoder (\textcolor{YellowOrange}{yellow} section in \cref{fig:method}) with two CA layers and two linear layers ($F(.)$) to generate online handwriting points sequentially. To generate the state ($\hat{p}^{w,i}_t$) of the $t$-th point, we concatenate the content feature $Z^c_i$ with the previous point states $\{p^{w,i}_j\}_{j=1}^{t-1}$, forming the content-conditioned point sequence $P^c_i=\{Z^c_i, p^{w,i}_1, ..., p^{w,i}_{t-1}\}$. Next, we apply a self-attention mechanism followed by layer normalization to obtain $\bar{P}^{c}_i$, which serves as the query vector. Finally, using cross-attention, we estimate $\hat{p}^{w,i}_t$, where $S^s_w$ and $G^s_w$ act as the ``key and value vectors'' of the first and second CA layer accordingly. This process is formalized as:
\begin{equation}
\begin{aligned}
Z^{cs}_{w,i} &= \text{CA}({\bar{P}}^{c}_i, {\bar{S}}^s_w) + P^{c}_i, \\
\hat{o}^{w,i}_t &= F\big(\text{CA}({\bar{Z}}^{cs}_{w,i}, {\bar{G}}^s_w) + Z^{cs}_{w,i}\big).
\end{aligned}
\vspace{-.2em}
\label{eq:5}
\end{equation}
In the 1st CA layer, $P^c_i$ is enriched with stroke-style information $S^s_w$, forming $Z^{cs}_{w,i}.$ In the 2nd CA layer, followed by $F(.)$, the glyph-style characteristics $G^s_w$ are further incorporated to generate the point location probability, modeled by a Gaussian mixture model, and three pen states, denoted as  $\hat{o}^{w,i}_t = \left[\{ \hat{\pi}^r, \hat{\mu}_x^r, \hat{\mu}_y^r, \hat{\delta}_x^r, \hat{\delta}_y^r, \hat{\rho}_{xy}^r \}_{r=1}^{R}, {\hat{m}}^1_t, {\hat{m}}^2_t, {\hat{m}}^3_t \right]$. The predicted output is then used to obtain the $t$-th point $\hat{p}^{w,i}_t = \{\Delta {\hat{x}}_t, \Delta {\hat{y}}_t, {\hat{m}}^1_t, {\hat{m}}^2_t, {\hat{m}}^3_t\}$ following \cite{SDT, WriteLikeYou}. By generating one point at a time, we obtain the online handwriting trajectory $\hat{Y}_{w,i} =\{\hat{p}_j^{w,i}\}_{j=0}^{N_p}$, representing the handwriting of the $w$-th writer for the $i$-th character. Our training strategy (\cref{sec:training}) ensures $\hat{Y}_{w,i}$ aligns with the ground truth $Y_{w,i}$.

\subsection{Training Strategy} \label{sec:training}

We optimize DNA in two stages: \textbf{stage 1} trains the entire model, while \textbf{stage 2} fine-tunes the adaptive style branch using the spacing loss $\mathcal{L}_{sp}$, keeping the content branch frozen to generate more natural and fluid samples.

\noindent\textbf{Baseline loss.} First, we apply the commonly used losses \cite{SDT} to train our DNA and define them as the baseline loss $\mathcal{L}_{base}$. Specifically, the negative log-likelihood loss $\mathcal{L}_{xy}$ and the cross-entropy loss $\mathcal{L}_{state}$ minimize the gap between DNA’s point state output $\hat{o}^{w,i}_t$ and the ground truth point $p_j^{w,i}$. Furthermore, to ensure the style encoder captures each writer’s unique handwriting style, the style loss $\mathcal{L}_{sty}$ reduces the distance between embeddings of the same writer (positive pairs) while increasing the distance between those of different writers (negative pairs). Similarly, the glyph loss $\mathcal{L}_{gly}$ promotes diverse representations by bringing patches of the same character closer (positive) while pushing patches from different characters apart (negative). The overall baseline loss function is then defined as:  
\begin{equation}  
\mathcal{L}_{base} = 2 \mathcal{L}_{state} + \mathcal{L}_{xy} + \mathcal{L}_{sty} + \mathcal{L}_{gly}.
\end{equation}

\begin{figure}[!t]
  \centering
    \includegraphics[width=0.7\columnwidth]{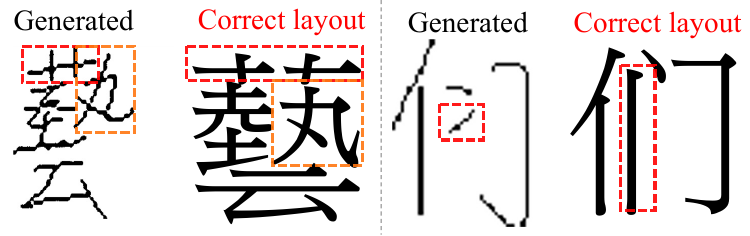}
    \vspace{-.5em}
    \caption{Stroke errors after \textbf{Stage 1} training are highlighted by bounding boxes in the `Generated' samples, while the `Correct layout' illustrates the desired stroke positions.
    }
    \vspace{-1.5em}
    \label{fig:wrong_pos}
\end{figure}

\noindent\textbf{Stage 1.} Beyond the baseline loss, we introduce the content loss $\mathcal{L}_{ct}$, including character loss $\mathcal{L}_{ch}$ and decomposition loss $\mathcal{L}_{de}$ to guide the adaptive content branch in focusing on character constituent components. Thus, it improves generalization on unseen characters with limited training data. The total loss function in stage 1 is shown as follows:
\begin{equation}
    \mathcal{L}_{stage 1} = \mathcal{L}_{base} + \lambda_1\mathcal{L}_{ct},
    \label{eq:stage1}
\end{equation}
where $\mathcal{L}_{ct} = \mathcal{L}_{ch} + \mathcal{L}_{de}$. To compute the cross-entropy loss $\mathcal{L}_{ch} =\sum^{N_{c}}_{i=1} -y^{ch}_i \log(\hat{y}^{ch}_i)$, the fusion features $Z^c_i$ are passed through a linear layer to predict $\hat{y}^{ch}_i$. Here, $y^{ch}\in \{0, 1\}^{N_c}$ represents a one-hot GT vector with $y^{ch}_i = 1$ at the $i$-th character, and $y_{j\neq
i} = 0$. $\hat{y}^{char}$ is the predicted probabilities over $N_c$ characters. Simultaneously, $Z^{lg}_i$ is processed through an LSTM layer followed by a linear layer to predict structure $\hat{st}^i$ and components $\hat{cp}^i$, which is used to compute decomposition ($de$) loss as follows:
\begin{equation}
\mathcal{L}_{de} = - \Big(\sum^{N_{st}}_{k=1} st^i_k \log(\hat{st}^i_k) + \sum^{N_{cp}}_{h=1} cp^i_h \log(\hat{cp}^i_h)\Big),
\vspace{-.2em}
\end{equation}
where $N_{st}$ and $N_{cp}$ are the number of structures and components of $i$-th character, respectively. $st^i_k$ and $cp^i_h$ represent the ground truth one-hot vectors for the $k$-th structure and $h$-th component with the length of $12$ and $9813$, respectively. By providing explicit guidance, the content branch can learn character components and structures accurately.

\noindent\textbf{Stage 2.} After training stage 1, we observed that the adaptive style branch exhibited inaccuracies in inter-stroke spacing, as illustrated in \cref{fig:wrong_pos}. To mitigate these gaps, we employ an additional spacing loss $\mathcal{L}_{sp}$ to fine-tune the style branch and decoder while the content branch is frozen. We denote bounding boxes (strokes) $\hat{S}_s$ and $S_s$ as the predicted and ground truth strokes, respectively, where stroke boundaries are determined by the pen-up state. Then, the spacing loss is designed to minimize the distance between $\hat{S}_s$ and $S_s$ as follows:
\begin{equation}
\mathcal{L}_{sp} = \frac{1}{N_{str}} \sum_{s=1}^{N_{str}} \left( 1 - \frac{|\hat{S}_s \cap S_s|}{|\hat{S}_s \cup S_s|} + \frac{|C_s \setminus (\hat{S}_s \cup S_s)|}{|C_s|} \right),
\end{equation}
where $N_{str}$ is the number of strokes. $|\hat{S}_s \cap S_s|$ represents the area of intersection between $\hat{S}_s$ and $S_s$, while $|\hat{S}_s \cup S_s|$ represents the area of their union. $C_i$ is the smallest bounding box enclosing both $\hat{S}_s$ and $S_s$. Thus, $|C_s \setminus (\hat{S}_s \cup S_s)|$ quantifies the non-overlapping area between $\hat{S}_s$ and $S_s$. Minimizing $\mathcal{L}_{sp}$ effectively reduces the inter-stroke distance for both overlapping and non-overlapping cases. The total loss function in stage 2 is defined as follows:
\begin{equation}
\mathcal{L}_{stage 2} = \mathcal{L}_{base} + \lambda_2\mathcal{L}_{sp}.
\label{eq:stage2}
\end{equation}

\section{Experimental Results} \label{sec:4_experi}

\subsection{Experiment Setup}
\label{sec:4_experi_setup}
\textbf{Dataset.} In this paper, we strive to handle the challenging Traditional Chinese dataset, SCUT-COUCH2009 \cite{SCUT-COUCH2009}, a dataset containing a total of $65$ writers, with each writer providing $5401$ characters. To facilitate evaluation in the unseen OHG task, we split the Traditional Chinese dataset into the SWSC, UWSC, and UWUC subsets. Specifically, SWSC serves as the training set, consisting of $55$ seen writers and $4,807$ seen characters. For evaluation, UWSC includes 10 unseen writers, each with $4,807$ seen characters (matching SWSC), and UWUC includes $10$ unseen writers with $594$ unseen characters per writer.
\textcolor{black}{Additionally, we use TUAT HANDS \cite{matsumoto2001collection} database for Japanese. The dataset includes approximately 3 million online Japanese handwritings, containing $4229$ characters and $271$ writers. We randomly select $2303$ common characters and $216$ writers as the SWSC setting and use them for training. For UWSC testing, the dataset consists of $2303$ seen characters from $55$ unseen writers. The remaining $1926$ unseen characters are used in UWUC testing.}

    

\noindent \textbf{Metrics.}  \label{sec:metrics} Consistent with prior handwriting generation research \cite{SDT, WriteLikeYou}, we employ Dynamic Time Warping (DTW) \cite{chen2022complex} to measure the distance between generated and ground truth handwriting trajectories. Additionally, we train content recognizers to compute the Content Score (CS) \cite{Deep-imitator}. The content recognizer is trained to classify characters across various writing styles, making it suitable for assessing whether a handwriting generator produces recognizable content. For online and offline generation settings, the content recognizer takes handwriting trajectories and images as inputs, respectively. \textcolor{black}{To assess human-like handwriting styles, we utilize FID and Geometry Score \cite{GS} (GS, in $\times10^{-3}$). Moreover, to demonstrate DNA's capability in enhancing the generalization of handwriting text recognition (HTR) models to unseen data, we train a HTR model \cite{HTR}, using a combination of real-seen and extra-generated samples. We evaluate its performance using the Character Error Rate (CER) metric.}



\noindent \textbf{Implementation details.} We fix the length of $\mathbf{v}^{compo}_{i}$ to $12$, using zero-padding to ensure compatibility with characters containing a maximum of $12$ components. Following previous OHG studies \cite{SDT, Deep-imitator}, we render offline style images $I^w_i$ from the online coordinate points of characters. For a fair comparison with prior works \cite{SDT, WriteLikeYou}, we set the number of grayscale style images to $N_s=15$ and resize them to $64 \times 64$. In line with prior works \cite{SDT, WriteLikeYou}, we adopt ResNet18 \cite{ResNet} and two Transformer layers \cite{attention} as the style encoder $E_s$ and the global encoder $E^g_c$. The decoder employs four attention layers. Additionally, the local encoder $E^l_c$ and the fusion decoder each use two attention layers. We optimize our DNA using Adam optimizer with a learning rate of $0.00002$ and a mini-batch size of $16$ on a single NVIDIA RTX 2080 GPU. The loss weights are set to $\lambda_1 = 0.5$ and $\lambda_2 = 1$. During testing, the maximum output sequence length $N_p$ is constrained to $300$.

\begin{table*}[!t]
\centering
\resizebox{\textwidth}{!}{
\begin{tabular}{@{\hspace{.1em}}c@{\hspace{.1em}}|@{\hspace{.3em}}c@{\hspace{.1em}}|@{\hspace{.1em}}c@{\hspace{.3em}}c@{\hspace{.3em}}c@{\hspace{.3em}}c@{\hspace{.3em}}c@{\hspace{.3em}}@{\hspace{.1em}}|@{\hspace{.3em}}c@{\hspace{.3em}}c@{\hspace{.3em}}c@{\hspace{.3em}}c@{\hspace{.3em}}c@{\hspace{.3em}}@{\hspace{.1em}}|@{\hspace{.3em}}c@{\hspace{.3em}}c@{\hspace{.3em}}c@{\hspace{.3em}}c@{\hspace{.3em}}c@{\hspace{.3em}}@{\hspace{.1em}}|@{\hspace{.3em}}c@{\hspace{.3em}}c@{\hspace{.3em}}c@{\hspace{.3em}}c@{\hspace{.3em}}c@{\hspace{.1em}}}
\toprule

\multirow{3}{*}{Type} & \multirow{3}{*}{Method} & \multicolumn{10}{c@{\hspace{.3em}}|@{\hspace{.3em}}}{Traditional Chinese} & \multicolumn{10}{c}{Japanese} \\

& & \multicolumn{5}{c@{\hspace{.3em}}|@{\hspace{.3em}}}{UWSC} & \multicolumn{5}{c@{\hspace{.3em}}|@{\hspace{.3em}}}{UWUC} & \multicolumn{5}{c@{\hspace{.3em}}|@{\hspace{.3em}}}{UWSC} & \multicolumn{5}{c}{UWUC} \\

& & DTW\textcolor{red}{$\downarrow$} 
& CS \textcolor{red}{$\uparrow$} 
& CER\textcolor{red}{$\downarrow$} 
& FID\textcolor{red}{$\downarrow$} 
& GS\textcolor{red}{$\downarrow$} 
& DTW\textcolor{red}{$\downarrow$} 
& CS \textcolor{red}{$\uparrow$} 
& CER\textcolor{red}{$\downarrow$} 
& FID\textcolor{red}{$\downarrow$} 
& GS\textcolor{red}{$\downarrow$} 
& DTW\textcolor{red}{$\downarrow$} 
& CS\textcolor{red}{$\uparrow$} 
& CER\textcolor{red}{$\downarrow$} 
& FID\textcolor{red}{$\downarrow$} 
& GS\textcolor{red}{$\downarrow$} 
& DTW\textcolor{red}{$\downarrow$} 
& CS\textcolor{red}{$\uparrow$} 
& CER\textcolor{red}{$\downarrow$} 
& FID\textcolor{red}{$\downarrow$} 
& GS\textcolor{red}{$\downarrow$} \\

\midrule
\multirow{3.2}{*}{\rotatebox{90}{Offline}}
& Dg-font \cite{Dg-font} & - & 8.63 & 1.39 & 40.00 & 0.57 & - & 10.86 & 3.09  & 42.07 & 2.29 & - & 72.76 & 2.32 & 39.81 & 31.89 & - & 18.08 & 15.26 & 39.48 & 1.62 \\
& Cf-font \cite{Cf-font} & - & 20.38 & 2.96 & 43.10 & \underline{0.15} & - & 18.28 &  3.82 & 44.88 & \textbf{1.05} & - & 71.20 & 2.34 & 41.43 & 15.73 & - & 18.41 & 15.25 & 41.09 & 1.92 \\
& FontDiffuser \cite{Fontdiffuser} & - & 82.72 & 1.51 & 41.64 & 7.40 & - & 78.65 & 2.97 & 45.09 & 2.45 & - & 90.84 & 1.35 & 55.01 & 63.27 & - & 50.08 & 15.24 & 60.21 & 3.35 \\

\midrule
\multirow{3.2}{*}{\rotatebox{90}{Online}}

& WriteLikeYou \cite{WriteLikeYou} & 1.1729 & 68.46 & \underline{1.33} & \textbf{9.86} & 0.91 & 3.7128 & 48.91 & 5.37 & 38.27 & 2.16 & \underline{0.9350} & 91.97 & 1.33 & 3.19 & \underline{4.41} & 0.9626 & \underline{51.17} & 9.72 & 4.81 & \underline{0.41} \\
& SDT \cite{SDT} & \textbf{1.0023} & \underline{83.54} & 1.97 & 13.77 & 0.20 & 1.2079 & \underline{81.04} & \underline{2.95} & \underline{16.43} & 2.67 & 1.0256 & \underline{92.38} & \underline{1.31} & \underline{1.49} & 7.29 & \underline{0.9216} & 50.57 & \underline{8.15} & \underline{2.16} & 0.45 \\

& DNA & \underline{1.1601} & \textbf{90.22} & \textbf{1.16} & \underline{11.99} & \textbf{0.07} & \textbf{1.0431} & \textbf{89.80} & \textbf{2.88} & \textbf{15.04} & \underline{1.50} & \textbf{0.8791} & \textbf{92.65} & \textbf{1.26} & \textbf{1.46} & \textbf{0.81} & \textbf{0.8675} & \textbf{55.09} & \textbf{6.80} & \textbf{1.73} & \textbf{0.23} \\
\bottomrule
\end{tabular}
}
\vspace{-.7em}
\caption{Quantitative comparison of offline and online methods on the Traditional Chinese and Japanese datasets under the UWSC and UWUC settings. The top and second-best accuracy results are highlighted.} 
\vspace{-1em}
\label{tab:compare_trad}
\end{table*}

\subsection{Comparing with SoTA Methods}
We conduct both quantitative and qualitative comparisons of our proposed method with previous offline methods, such as Dg-font \cite{Dg-font}, Cf-font \cite{Cf-font}, Fontdiffuser \cite{Fontdiffuser} and online works like WriteLikeYou \cite{WriteLikeYou}, and SDT \cite{SDT}.

\noindent \textbf{Quantitative comparison.} 
\textcolor{black}{As shown in \cref{tab:compare_trad}, our DNA method achieves SoTA performance on most metrics in Traditional Chinese. For demonstrating the capability of generating recognizable and reasonable characters, DNA achieves remarkable CS scores of $90.22$ and $89.80$ on UWSC and UWUC, respectively. For style imitation, it surpasses the second-best method with absolute gains of $0.07$ in GS (UWSC), $15.04$ in FID (UWUC). These results highlight DNA’s effectiveness in addressing unseen OHG. Furthermore, in the task of improving HTR models, samples generated by DNA improve the HTR model's baseline CER ({$1.62$}) in the UWSC setting and uniquely enable generalization to the UWUC setting. By prioritizing accurate character generation, DNA achieves superior HTR performance regarding CER. We also conduct the comparison on a Japanese handwriting dataset. As shown in \cref{tab:compare_trad}, DNA achieves SoTA results in both content and style evaluation.}


\begin{figure*}[!t]
\centering
\includegraphics[width=0.9\textwidth]{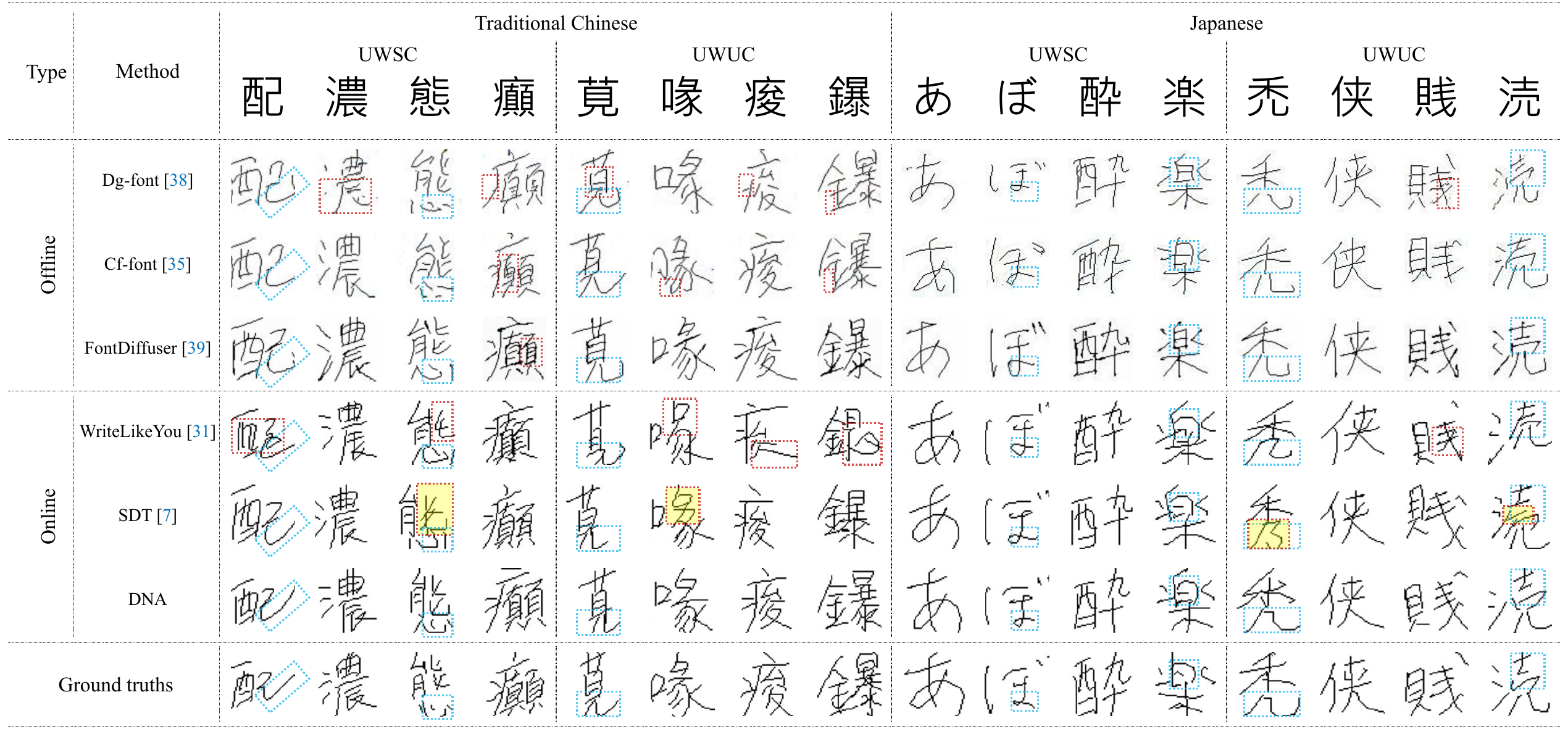}

\vspace{-.7em}
\caption{Qualitative comparison of synthesized handwriting from offline and online methods on the UWSC and UWUC testing sets. Red boxes mark areas where content is incorrect. Blue boxes outline the stylistic expression of some strokes.}
\label{fig:quanti_trad}
\vspace{-1.5em}
\end{figure*}

\noindent \textbf{Qualitative comparison.} \textcolor{black}{We examine both synthesized Traditional Chinese and Japanese character images generated by offline and online handwriting methods.} As observed in \cref{fig:quanti_trad}, offline methods, which rely on image-to-image translation, often overlook the important trajectory information inherent in written handwriting. The diffusion-based method, Fontdiffuser \cite{Fontdiffuser}, improves clarity and content correctness but still struggles to generate realistic handwriting. While online methods address handwriting flexibility. However, they face challenges in the \textcolor{black}{unseen OHG} task, e.g., SDT generates incorrect characters, as highlighted by the \textcolor{YellowOrange}{yellow} squares. In contrast, our DNA method generates more complete and structurally correct characters.

\vspace{-.5em}
\subsection{Analysis}
\noindent \textbf{Ablation studies.} Table \ref{tab:ablation} reports the ablation study of different loss combinations under UWSC and UWUC settings. Using only \(\mathcal{L}_{base}\) leads to the local and global encoders are not explicitly guided, which makes relatively low CS and high FID values. Then, incorporating \(\mathcal{L}_{ct}\) provides clear guidance between local and global information, allowing them to complement each other more effectively and archiving CS of $74.18$ and $79.71$ on UWSC and UWUC, respectively. Furthermore, as character extraction becomes more precise, style integration improves, leading to \textcolor{black}{an improvement in FID to $15.04$ on UWSC and $18.72$ on UWUC}. However, using only \(\mathcal{L}_{sp}\) and overlooking \(\mathcal{L}_{ct}\) like the first ablation lead the performance significant degradation. Additionally, as mentioned in \cref{sec:training}, \(\mathcal{L}_{sp}\) enhances the spatial information of the generated word. Therefore, the full-setting achieves \textcolor{black}{the best CS of $90.22$ and $89.80$, and FID of $11.99$ and $15.04$} on UWSC and UWUC, respectively. Moreover, the DTW score also achieved the best, with $1.1601$ on UWSC and $1.0431$ on UWUC. 

\begin{table}[!t]
\centering
\resizebox{\columnwidth}{!}{
\begin{tabular}{@{}c@{\hspace{.3em}}c@{\hspace{.3em}}c@{\hspace{.32em}}|ccc|ccc@{}}
\toprule
\multicolumn{3}{c@{\hspace{.2em}}|}{Settings} & \multicolumn{3}{c|}{UWSC} & \multicolumn{3}{c}{UWUC} \\
$\mathcal{L}_{base}$ & $\mathcal{L}_{ct}$ & $\mathcal{L}_{sp}$ & DTW\textcolor{red}{$\downarrow$} & CS \textcolor{red}{$\uparrow$} & FID \textcolor{red}{$\downarrow$} & DTW\textcolor{red}{$\downarrow$} & CS \textcolor{red}{$\uparrow$} & FID \textcolor{red}{$\downarrow$} \\
\midrule
\checkmark & & & 4.2936 & 20.33 & 40.86 & 3.1803 & 25.45 & 44.64  \\ 
\checkmark & \checkmark & & 1.5529 & 74.18 & 15.04 & 1.4392 & 79.71 & 18.72 \\
\checkmark & & \checkmark & 2.7041 & 20.07 & 15.46 & 2.5245 & 26.40 & 20.19  \\ 
\checkmark & \checkmark & \checkmark & 1.1601 & 90.22 & 11.99 & 1.0431 & 89.80 & 15.04 \\
\bottomrule
\end{tabular}
}
\vspace{-.7em}
\caption{Ablation study of different loss combinations under the UWSC and UWUC settings on the Traditional Chinese dataset.}
\label{tab:ablation}
\vspace{-1.5em}
\end{table}

\begin{table}[!t]
\centering
\resizebox{\columnwidth}{!}{
\begin{tabular}{@{}cc@{\hspace{.1em}}|ccc|ccc@{}}
\toprule
\multicolumn{2}{@{}c@{\hspace{.3em}}|}{Content branch} & \multicolumn{3}{c|}{UWSC} & \multicolumn{3}{c@{}}{UWUC} \\
Local & Global & DTW \textcolor{red}{$\downarrow$} & CS \textcolor{red}{$\uparrow$} & FID \textcolor{red}{$\downarrow$} & DTW \textcolor{red}{$\downarrow$} & CS \textcolor{red}{$\uparrow$} & FID \textcolor{red}{$\downarrow$} \\
\midrule   

\checkmark &  & 1.5009 & 87.24 & 13.36 & 1.7105 & 75.17 & 20.02 \\
& \checkmark & 1.0222 & 86.30 & 4.84 & 0.9782 & 86.65 & 7.46 \\
\checkmark & \checkmark & 1.1601 & 90.22 & 11.99 & 1.0431 & 89.80 & 15.04 \\


\bottomrule
\end{tabular}
}
\vspace{-.8em}
\caption{Quantitative results of DNA with different combinations of local and global encoders on the adaptive content branch.}
\label{tab:content}
\vspace{-1em}
\end{table}

\begin{figure}[!t]
    \centering
    \begin{subfigure}[t]{0.48\columnwidth}
        \includegraphics[width=95pt]{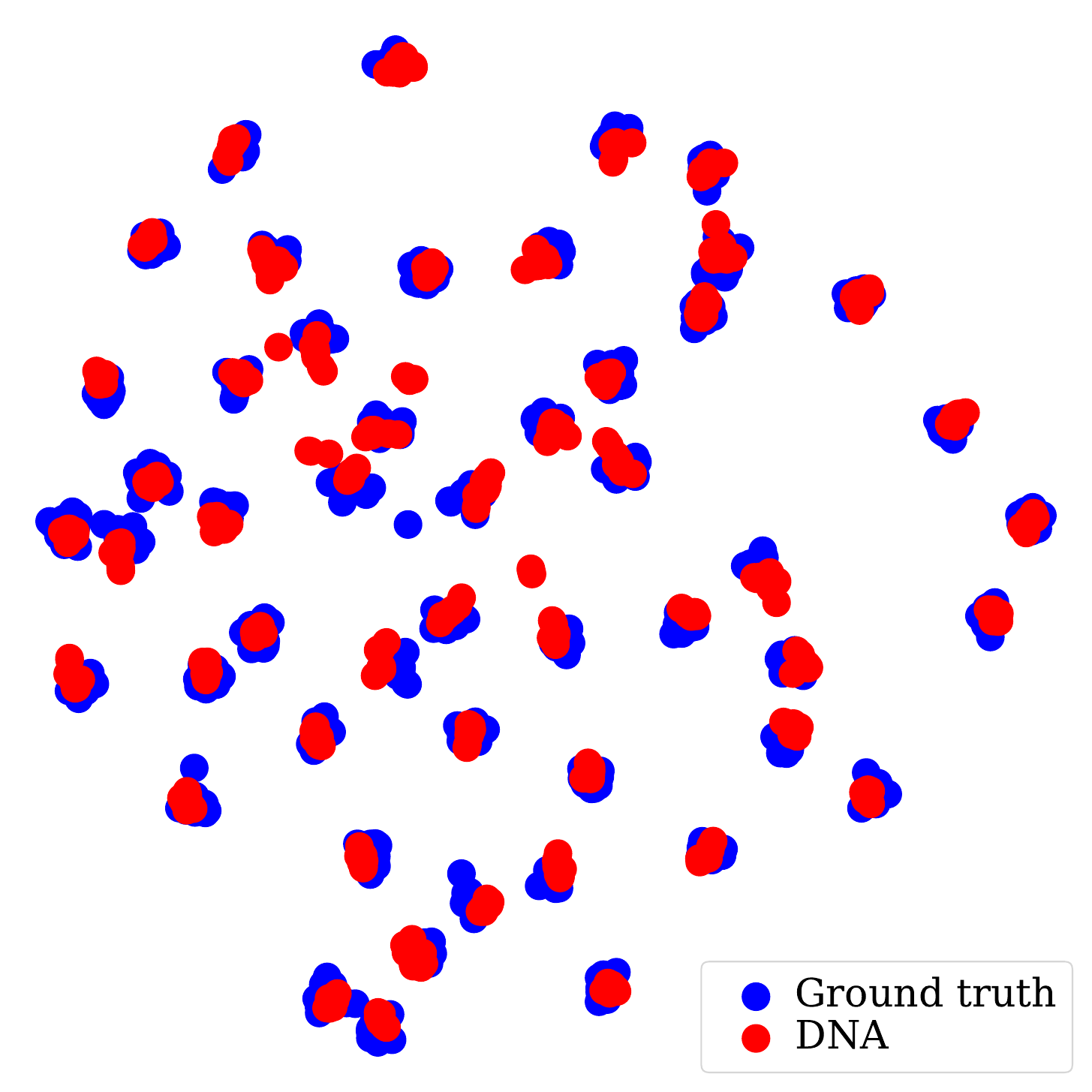}
        \caption{Content recognizer of UWSC.}
        \label{visualize_tsne_a}
    \end{subfigure}
    \hfill
    \begin{subfigure}[t]{0.46\columnwidth}
        \includegraphics[width=95pt]{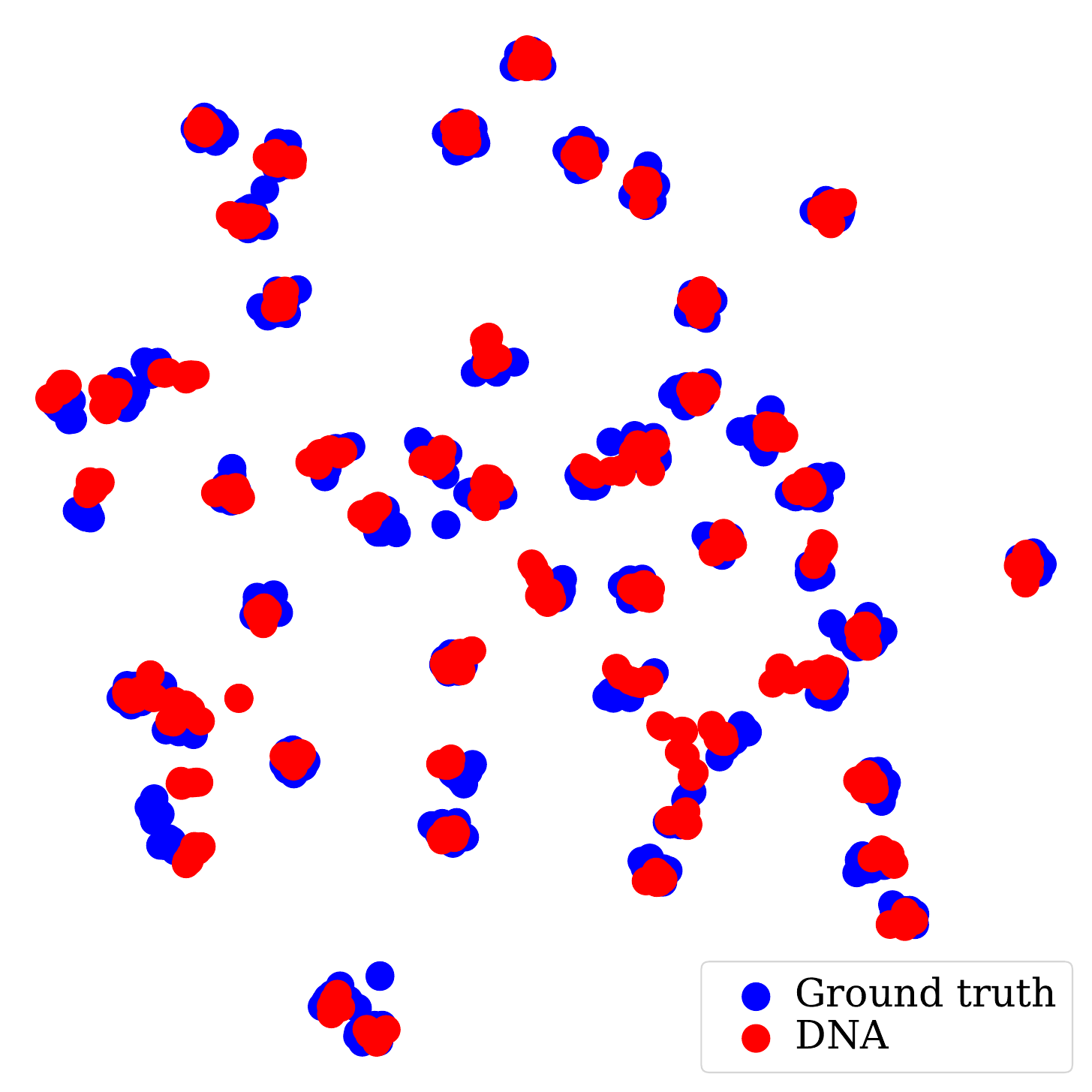}
        \caption{Content recognizer of UWUC.}
        \label{visualize_tsne_c}
    \end{subfigure}
    \vspace{-1em}
    \caption{t-SNE \cite{t-SNE} visualization for the content recognizers on the UWSC and UWUC datasets, with $50$ characters randomly selected. Ground truth features are represented by \textcolor{blue}{blue} markers, and DNA features by \textcolor{red}{red} markers.} 
    \vspace{-1.5em}
    \label{visualize_tsne} 
\end{figure}

\noindent\textbf{Local and global behavior.} To assess the impact of different encoder combinations on our model's performance, we evaluate the content branch based on both local and global information on Traditional Chinese, as reported in \cref{tab:content}. \textcolor{black}{As mentioned in \cref{sec:content}, the local encoder, which relies only on component combinations without direct character image input, is capable of generating structurally reasonable characters. The global encoder, on the other hand, has access to actual character images, leading to the best FID scores, since it can better preserve visual fidelity.
However, the global encoder alone does not always guarantee stroke-level accuracy, especially in fine-grained structural details. This is where the local encoder excels, maintaining competitive CS and DTW performance even without global guidance.
The full combination achieves the highest Content Score across both seen and unseen settings ($90.22$ on UWSC and $89.80$ on UWUC), indicating a more comprehensive understanding of content structure and style. This confirms that leveraging both sources of information is key to generating reasonable and high-quality characters.}

\noindent\textbf{Feature visualization.} We utilize t-SNE \cite{t-SNE} to examine the effectiveness of DNA in generating realistic handwriting through feature distribution analysis. Here, $50$ characters are randomly selected from both the UWSC and the UWUC datasets, respectively. As shown in Figs. \ref{visualize_tsne_a} and \ref{visualize_tsne_c}, DNA features exhibit strong alignment with ground truth features, demonstrating the adaptive content branch's ability to learn structural and texture information.

\begin{table}[!t]
\centering
\resizebox{\columnwidth}{!}{
\begin{tabular}{c|ccc|ccc}
\toprule
Style & \multicolumn{3}{c|}{UWSC} & \multicolumn{3}{c}{UWUC} \\

Samples & DTW \textcolor{red}{$\downarrow$} & CS \textcolor{red}{$\uparrow$} & FID \textcolor{red}{$\downarrow$} & DTW \textcolor{red}{$\downarrow$} & CS \textcolor{red}{$\uparrow$} & FID \textcolor{red}{$\downarrow$} \\
\midrule
30 & 1.1185 & 91.21 & 11.92 & 1.0007 & 90.52 & 15.04 \\
\rowcolor{gray!20} 15 & 1.1601 & 90.22 & 11.99 & 1.0431 & 89.80 & 15.04 \\
10 & 1.2345 & 89.38 & 12.23 & 1.1413 & 89.06 & 15.21 \\
5 & 1.3593  & 87.40 & 12.83 & 1.2011 & 87.04 & 15.73 \\
3 & 1.4892 & 84.92 & 13.36 & 1.3868 & 84.58 & 17.02 \\
1 & 2.1614 & 75.98 & 16.50 & 1.8042 & 75.52 & 20.95 \\
\bottomrule
\end{tabular}
}
\vspace{-.8em}
\caption{Performance of DNA with varying numbers of style samples during inference. The gray row highlights the state-of-the-art results achieved in the \textcolor{black}{unseen OHG} task.}
\label{tab:style}
\vspace{-1em}
\end{table}

\noindent\textbf{Effect of input style sample size.} To match real-world applications, where collecting sufficient style samples as the reference input is often challenging, we conducted experiments to assess the impact of varying the number of style samples $N_s$ from $1$ to $30$. As shown in \cref{tab:style}, $N_s=1$ and $N_s=30$ resulted in the worst and best performance, respectively, with FID scores of $16.50$ and $11.92$ for UWSC, and $20.95$ and $15.04$ for UWUC. Practically, collecting samples from $3$ to $15$ is feasible and yields satisfactory performance. For example, with only $15$ samples, DNA achieves superior results for the \textcolor{black}{unseen OHG} task, with DTW scores of $1.1601$ and $1.0431$, and CS scores of $90.22$ and $89.80$ for UWSC and UWUC, respectively.

\begin{table}[!t]
\centering
\resizebox{\columnwidth}{!}{
\begin{tabular}{@{\hspace{.3em}}l@{\hspace{.3em}}|M|M|M|M}
  \toprule
  \multirow{2}{*}{Method} & Model Size \textcolor{red}{$\downarrow$} &  Average speed \textcolor{red}{$\downarrow$} & Variance of speed \textcolor{red}{$\downarrow$} & FLOPs \textcolor{red}{$\downarrow$}\\
   & (MB) &  (sec/sample) & ($10^{-4}\times$sec/sample) & (in $\times10^{10}$) \\
  \midrule
  FontDiffuser~\cite{Fontdiffuser} & 383.36 & 1.69 & 171.41 & 69.60 \\
  Decoupling~\cite{ren2024decoupling} & 187.17 & 0.80 & 0.417 & 200.60 \\
  DNA & 279.11 & 0.13 & 17.657 & 60.53 \\
  \bottomrule
\end{tabular}
}
\vspace{-.8em}
\caption{Model size and inference speed of the diffusion-based method and our method DNA.}
\vspace{-1.5em}
\label{tab:speed}
\end{table}


\noindent \textbf{Diffusion models (DM) in the OHG task.} Diffusion-based methods have gained popularity and achieved impressive results in image generation. To assess their performance in OHG, we compare DNA with FontDiffuser~\cite{Fontdiffuser}, an offline DM handwriting generation method, and Decoupling~\cite{ren2024decoupling}, an online DM method. For fair efficiency comparisons, we only evaluate Decoupling’s character-generation branch without layout generation. As shown in \cref{tab:speed}, FontDiffuser has a model size $1.37 \times$ larger than DNA and an inference time about $13 \times$ slower per sample. In the online case, Decoupling requires roughly $3.3 \times$ more FLOPs than DNA. While diffusion models are powerful for generation tasks, DNA achieves competitive results with significantly lower inference cost, making it suitable for immediate applications. Although diffusion models demand more training and inference time currently, their potential remains high, suggesting a promising future direction of combining DNA with diffusion-based approaches.

\vspace{-1em}
\section{Conclusion}
\vspace{-.5em}
\label{sec:5_conclu} In this work, we address the challenges of handwriting generation under the unseen online handwriting generation (OHG) setting, aiming to generate both unseen characters and unseen handwriting styles during the testing phase. To tackle this, we propose a \underline{D}ual-branch \underline{N}etwork with \underline{A}daptation (DNA), consisting of an adaptive style branch and an adaptive content branch. The style branch captures stroke attributes such as writing direction, spacing, placement, and flow to generate realistic handwriting. Meanwhile, the content branch enhances generalization to unseen characters through local and global encoders. Experiments on the Traditional Chinese and Japanese datasets demonstrate that our method effectively generates human-like handwriting in the context of unseen OHG.

\section*{Acknowledgments}
This work was financially supported in part (project number: 112UA10019) by the Co-creation Platform of the Industry Academia Innovation School, NYCU, under the framework of the National Key Fields Industry-University Cooperation and Skilled Personnel Training Act, from the Ministry of Education (MOE) and industry partners in Taiwan.  It also supported in part by the National Science and Technology Council, Taiwan, under Grant NSTC-112-2221-E-A49-089-MY3, Grant NSTC-110-2221-E-A49-066-MY3, Grant NSTC-111-2634-F-A49-010, Grant NSTC-112-2425-H-A49-001, and in part by the Higher Education Sprout Project of the National Yang Ming Chiao Tung University and the Ministry of Education (MOE), Taiwan. We also would like to express our gratitude for the support from E.SUN Financial Holding Co Ltd, Professor Yi-Ren Yeh, E.SUN colleagues, Hsuan-Tung Liu, and et al.

{
    \small
    \bibliographystyle{ieeenat_fullname}
    \bibliography{main}
}

\end{document}


\maketitlesupplementary

In this supplementary material, we provide further details to enhance the understanding of our work. Specifically, we include:

\begin{enumerate}
\item   \textbf{Detailed Notations:} Comprehensive definitions of all mathematical symbols and notations used throughout the manuscript.
\item   \textbf{Qualitative Input Style Sample Size:} Additional qualitative comparisons demonstrating the impact of varying the number of input style samples on the generated handwriting.
\item   \textbf{Qualitative Loss Combination:} Additional qualitative comparisons of generated handwriting resulting from different combinations of loss functions, providing visual evidence of the effect of each loss component.
\item   \textbf{Qualitative Adaptive Content Branch:} Additional qualitative comparisons to examine the impact of local and global representations.
\item   \textbf{Style–Content Trade-off:} Analysis in the trade-off relationship between style and content quality.
\item   \textbf{Style-guided Synthesis:} User preference and qualitative results of style-guided synthesis.
\item   \textbf{Ability and Limitation on Cross-linguistic:} Further experiments on cross-linguistic tasks demonstrating the extent to which DNA can perform in an open-set setting.
\item   \textbf{Experiment on ProtoSnap benchmark:} Experiments on a small cuneiform sign dataset.
\item   \textbf{Pre-processing of Decomposed Vectors:} Detailed explanations of the pre-processing of transferring each character into its structural and component vectors.
\item   \textbf{Implementation Details of Evaluators:} Detailed explanations of the training strategy for the online content recognizer, offline content classification, and the style recognizer used in the main paper.
\end{enumerate}

Additionally, for reference, the link to the Microsoft JhengHei font is \url{https://learn.microsoft.com/zh-hk/typography/font-list/microsoft-jhenghei}, and the Traditional Chinese online dataset is available at \url{http://www.hcii-lab.net/data/scutcouch/}.

\section{Detailed Notations} 
To ensure facilitate readability throughout this paper, we provide a list of abbreviations and symbols in \cref{notation}.

\section{Qualitative Input Style Sample Size}
\begin{figure*}[!t]
  \centering
   \includegraphics[width=\textwidth]{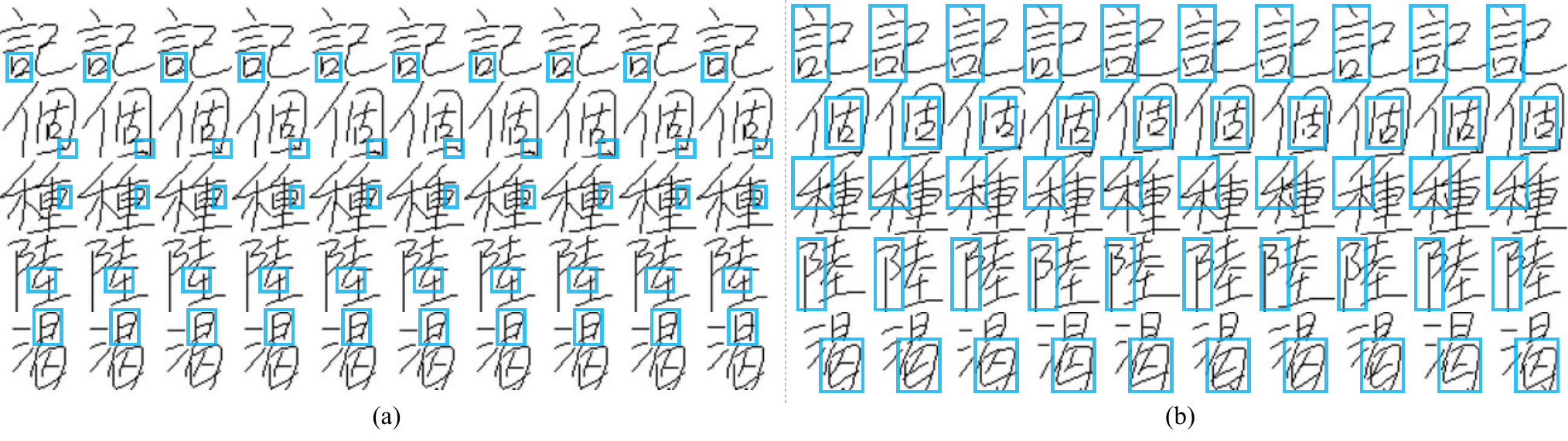}
   \vspace{-1.0em}
   \caption{(a) Generated results using $15$ style samples randomly selected from $16$ style samples. (b) Generated results using $15$ style samples randomly selected from $30$ style samples. The \textcolor{cyan}{blue} squares mark the subtle differences between each inference sample.}
   \label{fig:style}
\end{figure*}

\cref{fig:style} illustrates that by varying the content of style samples from the same writer, our model generates subtly different outputs. This variation aligns with real-life handwriting, where even when a person writes the same character multiple times, each instance will have slight differences. This flexibility allows our model to generate the natural diversity in handwriting styles, enhancing its realism.

\section{Qualitative Loss Combination} 
\begin{figure*}[!t]
  \centering
   \includegraphics[width=0.8\textwidth]{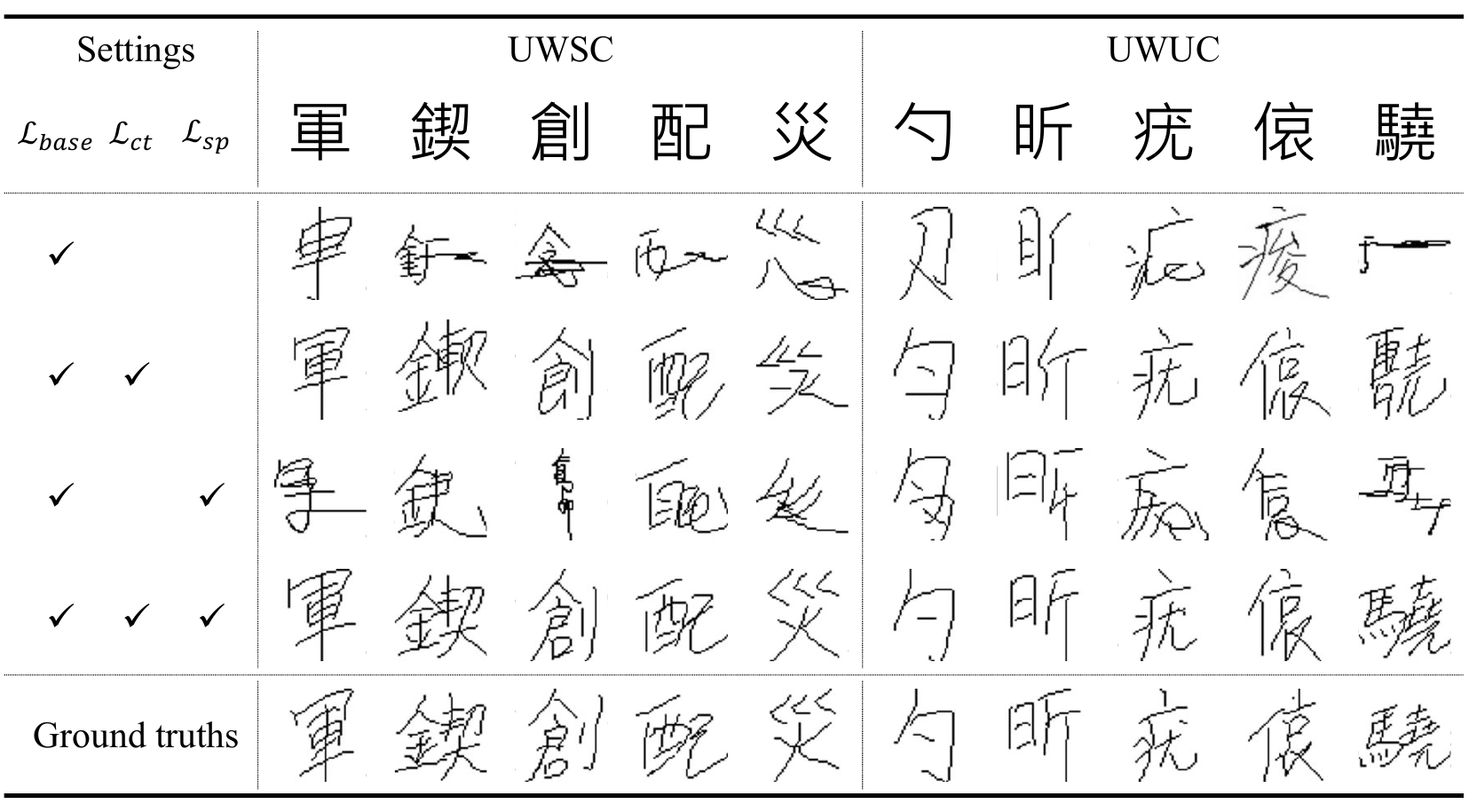}
   \vspace{-1em}
   \caption{Qualitative of DNA with different loss combinations under the UWSC and UWUC settings.}
   \label{fig:ablation}
\end{figure*}

\cref{fig:ablation} shows that overlooking $\mathcal{L}_{ct}$ causes the local and global encoders to lack clear guidance, leading to noise in the generated output. Applying $\mathcal{L}_{ct}$ helps define the character more clearly, making it closely resemble the target. However, stroke spacing remains inaccurate at both the character and writer levels. By incorporating the full setting, $\mathcal{L}_{sp}$ further refines the stroke spacing, resulting in a generated output that closely matches the target.

\section{Qualitative Adaptive Content Branch}
\begin{figure*}[!t]
  \centering

   \includegraphics[width=0.8\textwidth]{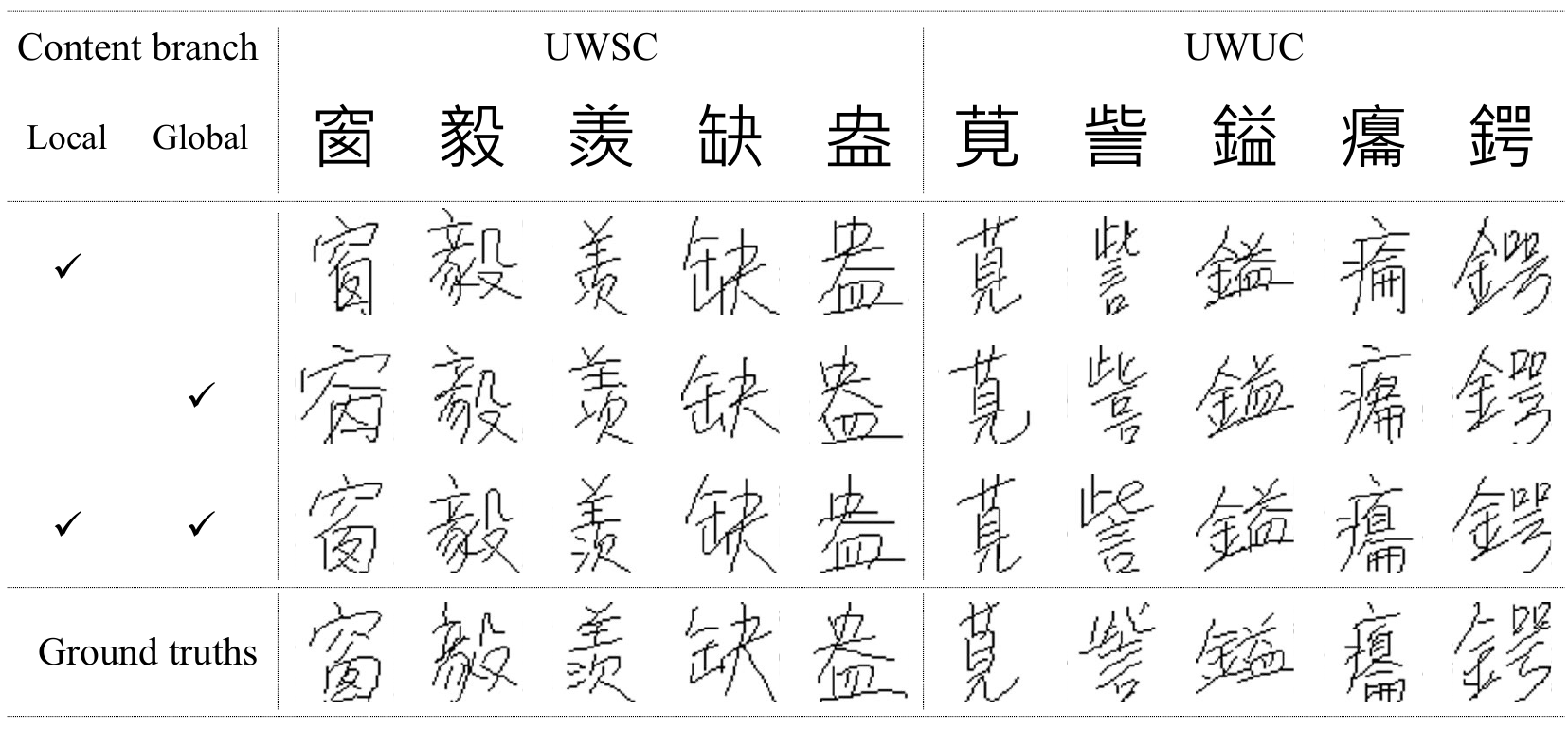}
   \vspace{-1em}
   \caption{ Qualitative of DNA with different combinations on the adaptive content branch under the UWSC and UWUC settings.}
   \label{fig:ablation2}
\end{figure*}

\textcolor{black}{\cref{fig:ablation2} illustrates that the local encoder effectively captures stroke-level details through component-based input, allowing the model to generate structurally correct characters even without seeing full images. In contrast, the global encoder, guided by actual character images, better preserves the overall character shape, but may generate less accurate stroke details. By combining both, the full model integrates fine-grained stroke control with global structural guidance, resulting in characters that are not only accurate and recognizable but also closely aligned with the target style.}

\begin{table*}[!htbp]
    \centering
    \begin{tabular}{ll}
    \toprule
    \textbf{Notation / Abbreviation} & \textbf{Description} \\
    \midrule

    \multicolumn{2}{l}{\textit{Overall Task and Settings}} \\
    \midrule
    OOHG & Open-set Online Handwriting Generation \\
    DNA  & Dual-branch Network with Adaptation \\
    SWSC & Seen-Writer-Seen-Characters \\
    UWSC & Unseen-Writer-Seen-Characters \\
    UWUC & Unseen-Writer-Unseen-Characters \\
    \midrule
    \multicolumn{2}{l}{\textit{Datasets, Inputs, and Outputs}} \\
    \midrule
    $X_s = \{x^s_w\}_{w=1}^{N_w}$ 
      & Style dataset of $N_w$ writers. \\
    $x^s_w = \{I^{w}_0, \dots, I^{w}_{N_s-1}\}$ 
      & The set of $N_s$ style images of the $w$-th writer. \\
    $X_c = \{(\mathbf{v}^c_i, I^c_i)\}_{i=1}^{N_c}$ 
      & Content dataset of $N_c$ characters. \\
    $\mathbf{v}^c_i = \{\mathbf{v}^{struct}_{i}, \mathbf{v}^{compo}_{i}\}$ 
      & Decomposition vectors for the $i$-th character (structure and components). \\
    $I^c_i \in \mathbb{R}^{3 \times H \times W}$ 
      & Printed character image for the $i$-th character. \\
    $Y_{w,i} = \{p_j^{w,i}\}_{j=1}^{N_p}$ 
      & Ground-truth online handwriting (point set) for writer $w$, character $i$. \\
    $p_j^{w,i} = \{\Delta x_j, \Delta y_j, m^1_j, m^2_j, m^3_j\}$ 
      & Each point’s offsets $(\Delta x_j,\Delta y_j)$ and pen states $m^1_j,m^2_j,m^3_j$. \\
    \midrule
    \multicolumn{2}{l}{\textit{Style Branch and Content Branch}} \\
    \midrule
    $E_s(\cdot)$, $h_s(\cdot)$, $h_g(\cdot)$ 
      & Style encoder, style head, and glyph head of the style branch. \\
    $Z^s_w = E_s(x^s_w)$ 
      & Extracted style features from writer $w$’s style images. \\
    $S^s_w = h_s(Z^s_w)$ 
      & Writer-specific style $(S^s_w)$ features. \\
    $G^s_w = h_g(Z^s_w)$ 
    & Glyph-specific style $(G^s_w)$ features.
    \\
    $E^l_c(\cdot)$ 
      & Local (structural) encoder. \\
    $E^g_c(\cdot)$ & Global (texture) content encoder. \\
    $Z^{lc}_i = E^l_c(\mathbf{v}^c_i)$ 
      & Local representation (structure + component) for character $i$. \\
    $Z^{gc}_i = E^g_c(I^c_i)$ 
      & Global texture representation for character $i$. \\
    $Z^c_i$ 
      & Fused content feature after cross-attention between $Z^{lc}_i$ and $Z^{gc}_i$. \\
    \midrule
    \multicolumn{2}{l}{\textit{Decoder and Predictions}} \\
    \midrule
    $\hat{p}^{w,i}_t$ 
      & Predicted point at step $t$ for writer $w$, character $i$. \\
    $\hat{Y}_{w,i} = \{\hat{p}^{w,i}_t\}_{t=0}^{N_p}$ 
      & Generated online handwriting sequence. \\
    \midrule
    \multicolumn{2}{l}{\textit{Loss Functions}} \\
    \midrule
    $\mathcal{L}_{xy}, \mathcal{L}_{state}$ 
      & Negative log-likelihood and cross-entropy losses for point prediction. \\
    $\mathcal{L}_{sty}, \mathcal{L}_{gly}$ 
      & Style and glyph contrastive losses. \\
    $\mathcal{L}_{ch}, \mathcal{L}_{de}$ 
      & Character classification and decomposition losses (part of $\mathcal{L}_{ct}$). \\
    $\mathcal{L}_{ct} = \mathcal{L}_{ch} + \mathcal{L}_{de}$ 
      & Content loss. \\
    $\mathcal{L}_{sp}$ 
      & Spacing loss. \\
    $\mathcal{L}_{base} 
       = 2\mathcal{L}_{state} 
       + \mathcal{L}_{xy} 
       + \mathcal{L}_{sty} 
       + \mathcal{L}_{gly}$ 
      & Baseline loss. \\
    \bottomrule
    \end{tabular}
    \caption{Notations and abbreviations}
    \label{notation}
\end{table*}

\begin{figure*}[!t]
\centering
\centering
\includegraphics[width=0.6\textwidth]{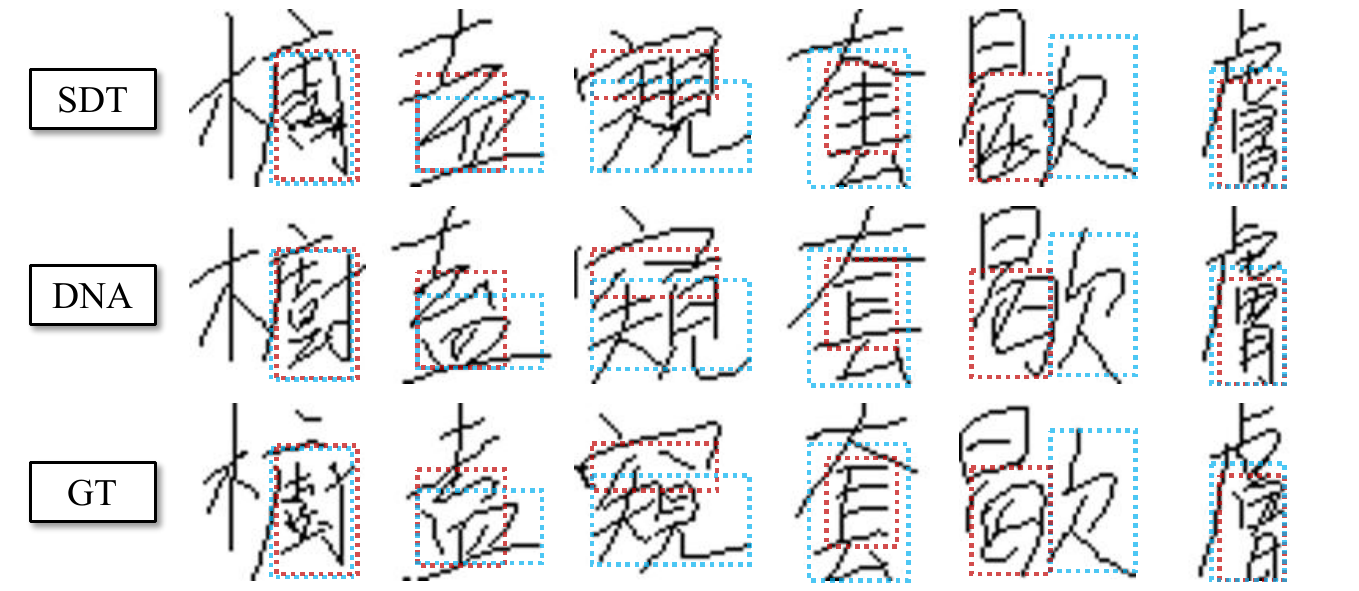}
\caption{Qualitative results highlighting the trade-off between style and content scores. DNA seeks a balanced outcome between style and content, whereas SDT tends to sacrifice content accuracy to improve the style score. The \textcolor{red}{red} and \textcolor{cyan}{blue} boxes indicate regions of interest for evaluating content correctness and style similarity to the ground truth (GT), respectively.}
\label{fig:eval-a}
\vspace{-1.em}
\end{figure*}

\section{Style–Content Trade-off}
Style and content performance are inherently interdependent: mimicking a ``scrawled'' style improves handwriting imitation but often harms recognizability, and vice versa. As shown in \cref{fig:eval-a} (\textcolor{cyan}{blue} squares), in examples (1, 2, and 6, from left to right), SDT closely mirrors the global style of the ground truth (GT), producing strokes that approximate the GT well and achieving a higher style score. However, this comes at the cost of structural precision, resulting in incorrect content. In contrast, DNA achieves a better balance between style and content, as highlighted in the (\textcolor{red}{red} squares), which facilitates accurate character generation. This improvement stems from the design of the content input: SDT relies solely on global character images, which may overlook fine-grained structural details, whereas DNA integrates both local component information and global content images, leading to better preservation of stroke details and local structure — ultimately producing results that are more faithful to both the local style and content of the GT. Nevertheless, this can result in a slight reduction in style score.

\begin{table}[!t]
\centering
\resizebox{0.6\columnwidth}{!}{
\begin{tabular}{c|c|c}
\hline

\multirow{2}{*}{Method} & \multicolumn{2}{c}{User Study (\%)} \\

& UWSC & UWUC \\

\hline
FontDiffuser & 24.78 & 18.23 \\
WriteLikeYou & 2.61 & 1.47 \\
SDT          & \underline{25.22} & \underline{27.06} \\

DNA          & \textbf{47.39} & \textbf{53.24} \\
\hline
\end{tabular}}
\caption{Handwriting generation results evaluated by user preference, showing the percentage of volunteers who judged the generated samples as more similar to the ground truth.}
\label{tab:user}
\end{table}

\begin{figure}[!t]
  \centering

   \includegraphics[width=\columnwidth]{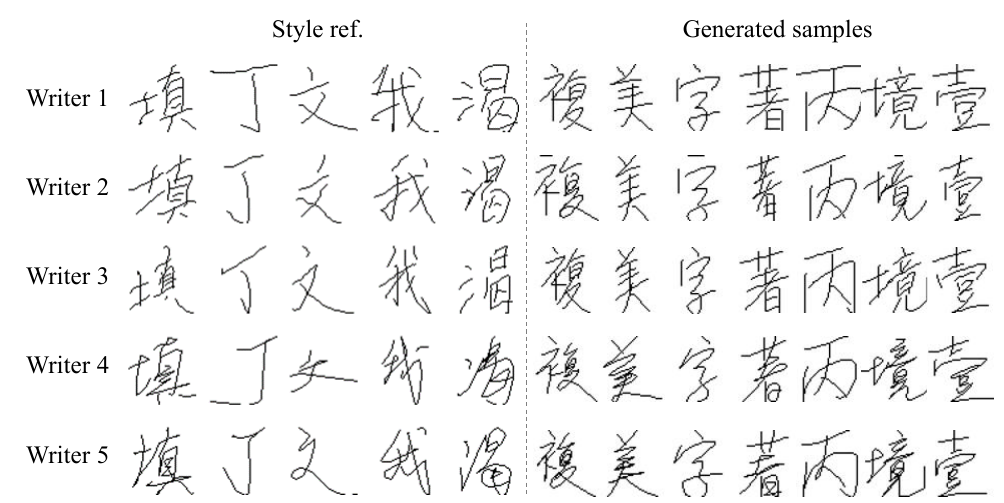}
   \vspace{-1em}
   \caption{Qualitative results of style-guided synthesis. The left side displays style reference inputs, while the right side shows the generated samples in the writers' style.}
   \label{fig:style_guide}
\end{figure}

\section{Style-guided Synthesis}
In \cref{tab:user}, we conduct a user study involving $50$ volunteers for voting on which generated samples are most similar to the ground truths. We received $47\%$ and $53\%$ user preference in the UWSC and UWUC settings, respectively. Additionally, we provide qualitative results for style-guided synthesis in \cref{fig:style_guide}. When the style references come from different writers, our model successfully generates other characters that faithfully reflect each writer’s unique handwriting style. Notably, for the fourth and fifth writers, whose references are drawn from another dataset characterized by more cursive handwriting, our method still captures the stylistic traits and produces characters that closely resemble the target writing style.

\begin{figure}[!t]
  \centering
    \includegraphics[width=\columnwidth]{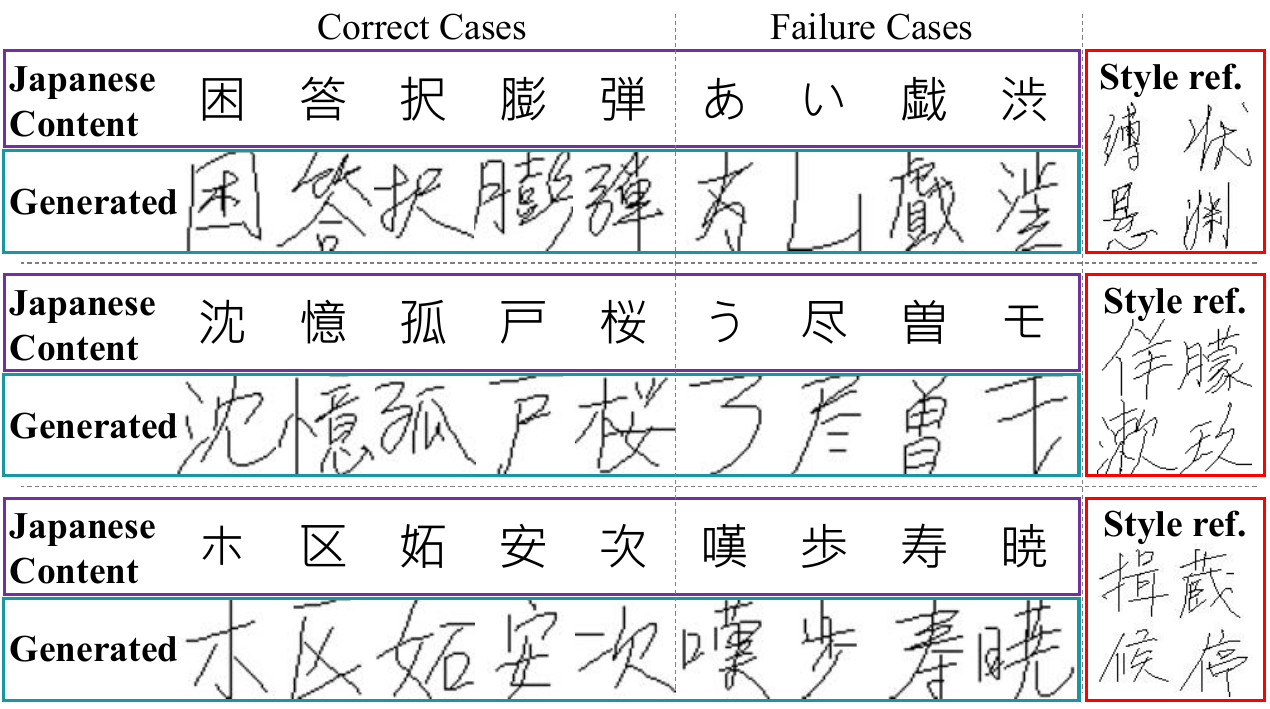}
    \vspace{-1.5em}
    \caption{Generalization of cross-language, where DNA is trained on Chinese characters and directly generates Japanese characters. The rightmost column shows the style reference.}
    \label{fig:cross}
\end{figure}

\begin{figure}[!t]
\centering
\includegraphics[width=\columnwidth]{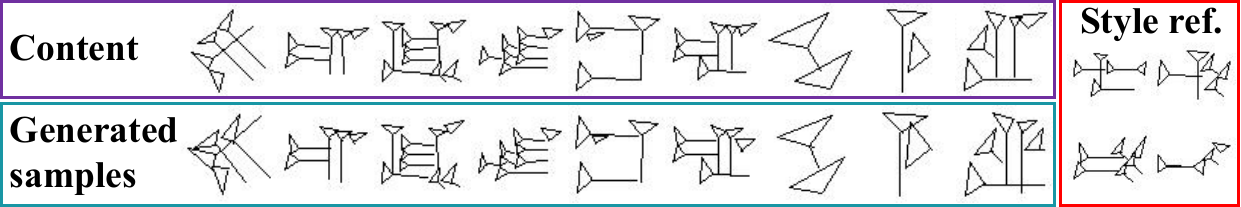}
\vspace{-1.8em}
\caption{Generalization of the ProtoSnap benchmark \cite{protosnap}, where DNA is well-trained on Chinese characters and fine-tuned on cuneiform signs. The rightmost column shows the style reference.}
\label{fig:protosnap}
\vspace{-1.em}
\end{figure}

\begin{figure*}[!t]
  \centering
   \includegraphics[width=0.8\textwidth]{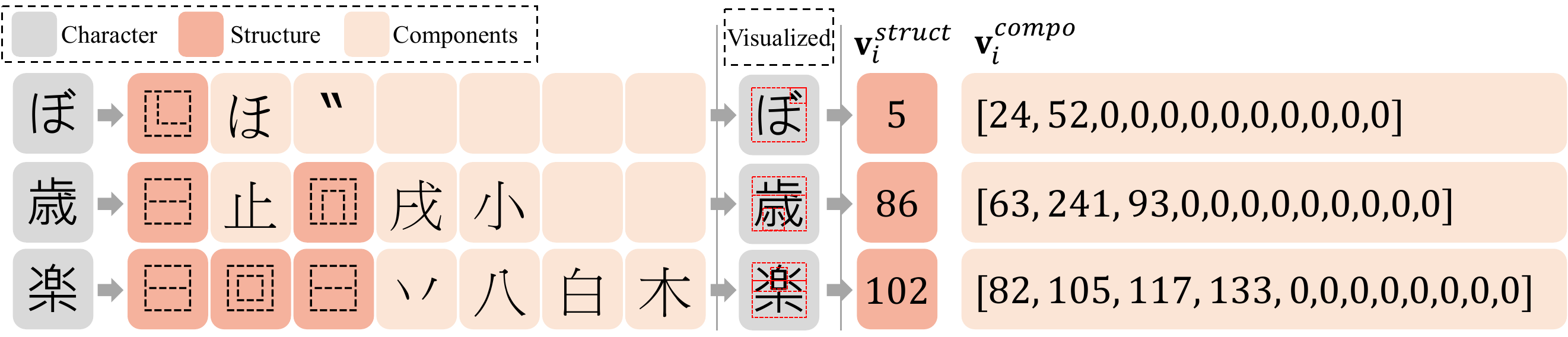}
   \vspace{-1em}
   \caption{Examples of decomposition of Japanese characters and their corresponding encoding vectors, $\mathbf{v}^{struct}_{i}$ and $\mathbf{v}^{compo}_{i}$.}
   \label{fig:compo_j}
\end{figure*}

\begin{figure}[!t]
  \centering
   \includegraphics[width=0.9\linewidth]{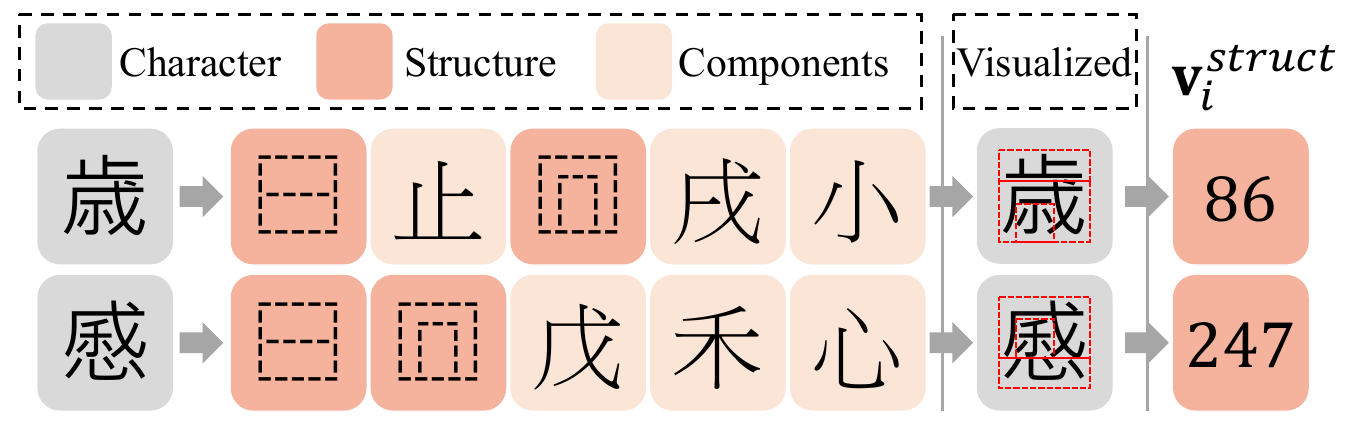}
   \vspace{-1em}
   \caption{Examples of different structural vectors from similar structure decompositions.}
   \label{fig:compo_struct}
\end{figure}

\section{Ability and Limitation on Cross-linguistic}
Cross-linguistic handwriting generation is challenging because character components differ across languages. Our method synthesizes unseen characters by recombining known components; however, generating characters composed entirely of unseen components remains difficult. In \cref{fig:cross}, we evaluate our model trained on traditional Chinese for generating Japanese characters. As shown on the left, characters sharing components with Chinese remain recognizable, whereas those composed of entirely novel components result in failed generations.

\section{Experiment on ProtoSnap benchmark}
We fine-tune DNA, originally well-trained on Chinese, using the ProtoSnap benchmark \cite{protosnap}, which provides prototype alignment for cuneiform signs. For fine-tuning and evaluation, we adopt the Assurbanipal subset of ProtoSnap, which contains 161 valid characters, with 131 used for training and 30 reserved for testing. In the case of cuneiform signs, the decompositions are straightforward since their components are explicitly reflected in the trajectory data, making it possible to identify both the components and structural organization of each character. This property makes cuneiform signs particularly suitable for DNA to naturally leverage component-structure alignments to enhance generation quality. The generated samples are shown in \cref{fig:protosnap}.

\section{Pre-processing of Decomposed Vectors}
Following the IDS (Ideographic Description Sequence) rules, both Chinese and Japanese characters can be decomposed into structures and components, as illustrated in Fig. \textcolor{wacvblue}{3} of the main paper and \cref{fig:compo_j}, respectively. We separate them to construct a structural vector and a component vector.

\noindent \textbf{Structural vector.} This vector is determined by all structural elements (dark orange) of a character and their relative positions. For example, as shown in \cref{fig:compo_struct}, the structure \includegraphics[width=0.3cm, height=0.3cm]{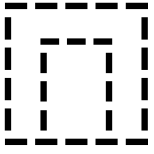} in the second position differs from \includegraphics[width=0.3cm, height=0.3cm]{fig/struct1.pdf} in the third position, and thus their structural indices are assigned differently to reflect their distinct roles in the character composition.

\noindent \textbf{Component vector.} Each component element corresponds to a unique component index, and the component vector is constructed by sequentially arranging these indices. Zero-padding is applied where necessary.


\section{Implementation Details of Evaluators} We split the UWSC dataset into $80$\% for training and $20$\% for testing. Similarly, the UWUC dataset is divided using the same $8$:$2$ ratio. Content evaluators are then trained on each dataset setting, and the trained evaluators are used to assess each set separately.

\begin{enumerate}
    \item \textbf{Dynamic Time Warping (DTW)} \cite{berndt1994using, chen2022complex}: We employ DTW as an elastic matching method to align two sequences, calculating the distance between the generated and actual characters. This alignment-based approach allows for a robust comparison of sequence similarity between generated and real characters.
    \item \textbf{Content Score (CS)} \cite{Deep-imitator}: To assess structural accuracy, we utilize a content recognizer trained specifically on the dataset’s training set to evaluate how accurately the generated character structures match the intended characters. The recognizer uses the Adam optimizer with a learning rate of $0.001$ and a batch size of $128$. Additionally, for offline methods, we use images as input for training, whereas for online methods, we use a coordinated sequence of characters for training.
    \item \textbf{Fréchet Inception Distance (FID)}: To quantitatively assess the stylistic similarity between generated and real handwriting, we use FID. This metric compares the distributions of high-level visual features, capturing both the realism and stylistic coherence of the generated handwriting.
    \item \textbf{Geometry Score (GS)} \cite{GS}: To complement the FID-based evaluation, we use GS to assess the structural of generated handwriting. This metric compares the topological properties of the generated and real data manifolds, revealing how well the model captures the underlying geometry of the handwriting style.

    \item \textbf{Character Error Rate (CER)}: To assess the benefit of generated data for downstream handwriting recognition, we trained a Handwritten Text Recognition (HTR) model \cite{HTR} with a combination of \colorbox{green!15}{real} and \colorbox{red!15}{generated} samples. The \colorbox{green!15}{real} set comes from the \textbf{SWSC} setting and includes all $4807$ characters from each of $55$ writers. For the \colorbox{red!15}{generated} data, two settings were used: in \textbf{UWSC}, we synthesized $4807$ characters for each of $10$ writers ($48070$ samples in total) and evaluated the model on the \textbf{UWSC} ground truth; in \textbf{UWUC}, we generated $594$ unseen characters for each of $10$ unseen writers ($5940$ samples) and tested on the corresponding ground-truth set.

\end{enumerate}

{
    \small
    \bibliographystyle{ieeenat_fullname}
    \bibliography{main}
}